# Image interpretation by iterative bottom-up top-down processing

**Shimon Ullman, Liav Assif, Alona Strugatski, Ben-Zion Vatashsky, Hila Levy, Aviv Netanyahu, Adam Yaari**

Department of Computer Science and Applied Mathematics, Weizmann Institute of Science, Israel

shimon.ullman@weizmann.ac.il, liav.assif@weizmann.ac.il , alona.faktor@weizmann.ac.il, ben-zion.vatashsky@weizmann.ac.il, hila.levi@weizmann.ac.il, avivn@mit.edu, yaari@mit.edu

## Abstract

Scene understanding requires the extraction and representation of scene components, such as objects and their parts, people, and places, together with their individual properties, as well as relations and interactions between them. We describe a model in which meaningful scene structures are extracted from the image by an iterative process, combining bottom-up (BU) and top-down (TD) networks, interacting through a symmetric bi-directional communication between them ('counter-streams' structure). The BU-TD model extracts and recognizes scene constituents with their selected properties and relations, and uses them to describe and understand the image.

The scene representation is constructed by the iterative use of three components. The first model component is a bottom-up stream that extracts selected scene elements, properties and relations. The second component ('cognitive augmentation') augments the extracted visual representation based on relevant non-visual stored representations. It also provides input to the third component, the top-down stream, in the form of a TD instruction, instructing the model what task to perform next. The top-down stream then guides the BU visual stream to perform the selected task in the next cycle. During this process, the visual representations extracted from the image can be combined with relevant non-visual representations, so that the final scene representation is based on both visual information extracted from the scene and relevant stored knowledge of the world.

We show how the BU-TD model composes complex visual tasks from sequences of steps, invoked by individual TD instructions. In particular, we describe how a sequence of TD-instructions is used to extract from the scene structures of interest, including an algorithm to automatically select the next TD-instruction in the sequence. The selection of TD instruction depends in general on the goal, the image, and on information already extracted from the image in previous steps. The TD-instructions sequence is therefore not a fixed sequence determined at the start, but an evolving program (or 'visual routine') that depends on the goal and the image.

The extraction process is shown to have favourable properties in terms of combinatorial generalization, generalizing well to novel scene structures and new combinations of objects, properties and relations not seen during training. Finally, we compare the model with relevant aspects of the human vision, and suggest directions for using the BU-TD scheme for integrating visual and cognitive components in the process of scene understanding.

**Contents**

# 1. General background and goals: combining vision and cognition

Vision allows us to understand the world around us based on the images reaching our eyes. Such understanding requires the extraction and representation of various scene components, including objects, people, and places, together with their individual properties, as well relations and interactions between them. We describe below a model that extracts and recognizes scene components with their properties and relations, and uses them to describe and understand the image.

In the general approach described below we propose that meaningful information is extracted from the image by an iterative process, combining bottom-up (abbreviated BU) and top-down (TD) streams, illustrated schematically in Figure 1 (Vision-cognition: Combining vision and cognition). In this scheme, the visual part of the model (labeled 'vision') constructs a scene representation at a higher-level, which we refer to as the cognitive part of the model (labeled 'Cognition'). The scene representation is constructed by the iterative use of three components, illustrated schematically in Figure 1. The first component, labeled 'visual stream', goes from vision to cognition, the second, labeled 'cognitive augmentation' takes place within the cognitive part, and the third component, labeled 'top-down stream' goes back from the cognitive to the visual part. Briefly, the visual stream delivers high-level, useful visual representations to the cognitive part of the system. The 'cognitive augmentation' component augments the extracted visual representation based on relevant non-visual stored representations. It also provides input to the top-down stream, which selects the relevant information to extract next. The top-down stream then guides the visual stream to extract the selected information in the next cycle. During this process, the visual representations extracted from the image are combined with relevant cognitive representations, so that the final scene representation is based on both visual information extracted from the scene and relevant stored knowledge of the world.

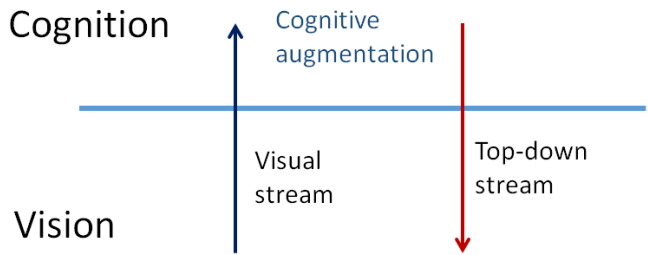

*Figure 1: Vision-cognition: Combining vision and cognition* A scene representation is constructed by the iterative use of three components: a 'visual stream', going from vision to cognition, 'cognitive augmentation' taking place within the cognitive part, and a 'top-down stream', going back from the cognitive to the visual part.

The roles of the three components and their integration is illustrated schematically using an image example shown in Figure 2a (a person stealing a purse). An example scene representation, which includes relevant scene components (e.g. man, woman, bag, chair), some of the properties (e.g. red bag) and relations (e.g. man grabbing a purse), is illustrated by the graph in Figure 2b (blue part). An important relation in this scene is that the sitting woman is the likely owner of the red purse. 'Ownership' is a semantic rather than direct visual relation, which often depends on specific knowledge; (e.g., that, unlike most cases, a porter, or a security guard, do not own the bag they manipulate). In the approach above, the proposed 'cognitive augmentation' part of the model will add relations and attributes based on relevant non-visual cognitive representation. Specifically, in the current scene, it will be able to use relevant stored knowledge to add the ownership relation between the sitting woman and the bag (dashed red in Figure 2b). The subsequent stage will be to use the top-down component, instructing the model to extract next the woman's direction of gaze (solid red). This will determine whether the person grabbing the bag is within her field of view, leading as a result to the conclusion regarding a likely 'stealing' interpretation.

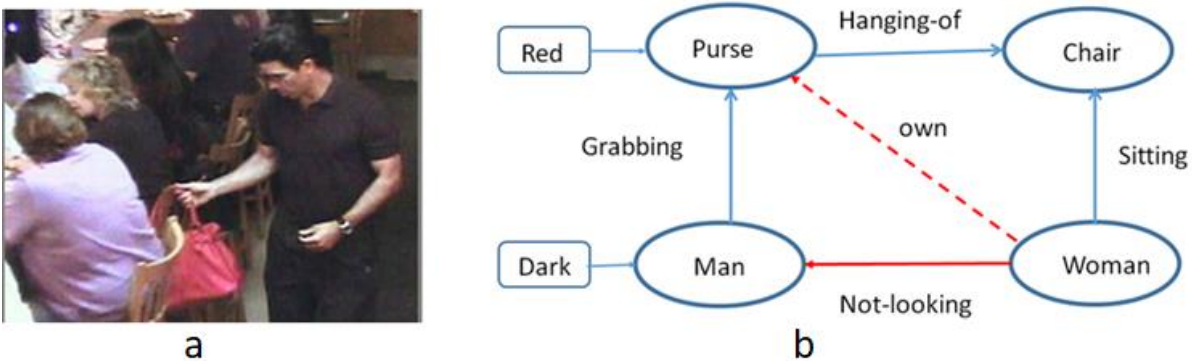

*Figure 2: Stealing* A man stealing a purse from a sitting woman. a. The input image. b. A scene description in terms of scene components, their properties and relations. Dashed red: added by cognitive augmentation. Solid red: extracted by the final top-down instruction.

We used the image (Figure 2a) along with other scenes in an on-line human study, which tracked the extraction of structured scene information (objects, people, properties, relations) from images over time (50 milliseconds – 2 seconds). The results of this psychophysical timeline study will be described elsewhere, for more details see Supplementary S1 (The psychophysical timeline study). For the image in Figure 2 (Stealing), the descriptions in (b) is a graphical summary of all the scene components, properties and relations reported by the majority of subjects by the time a 'stealing' event was reported. The extraction of multiple scene components and their inter-relations showed a process that developed gradually over time. For example, at a short presentation time of 100 milliseconds (msecs), only the man and woman were reported, the bag (125 msecs), 'sitting' and 'chair' (200 msecs) appeared later, and 'stealing' took longer than 500 msecs presentation time to be reported. The empirical time-line of the scene description supports the temporal order described above e.g., the late appearance of ownership relation, followed by the extraction of gaze direction. (See Figure 30 (Stealing timeline), in Supplementary S1.)

Within this general approach and three model components, the visual stream has been studied extensively in the past, both empirically and computationally. The cognitive augmentation part is a primarily non-visual and will not be discussed here. The focus of the paper will be on the bi-directional communication between the visual and cognitive components, combining the bottom-up and top-down streams in a so-called counter-streams structure. The next section describe the structure of the BU-TD network as well as its training procedure.

## 2. The BU-TD counter-streams network

### 2.1 Network structure
We describe first the general structure of the counter-stream model, and then describe the learning procedure used in training the model and the use of TD instructions to guide the visual process. The scheme is based on two network streams that provide bi-directional communication between the visual and cognitive components: a bottom-up (BU) and a top-down (TD) streams. The two are identical in structure, but directed in opposite directions, creating a 'counter-streams' structure, which captures aspects of cortical connectivity along the hierarchy of visual areas (Ullman, 1995; Markov *et al.*, 2013, 2014) The BU stream is modeled as a standard deep net (using, e.g. ResNet, (He *et al.*, 2016)). A schematic representation of the counter-streams structure is shown in Figure 3 (Counter-streams structure) below; a fuller description is given subsequently in Figure 5 (Full-structure).

Figure 3a illustrates a small part of the BU net (in green) and the complementary TD net (red). As shown, for each unit $x$ on the bottom-up stream there is a corresponding unit $\bar{x}$ on the top-down stream, and if $x$ is connected to a target unit $y$, the opposite connection from $\bar{y}$ to $\bar{x}$ exists on the top-down stream. Corresponding layers on the opposite streams are connected, in both directions, by cross-streams connections (not shown). More details of the structure are described in Figure 5 (Full-structure) below and in Supplementary S2 (The BU-TD structure).

As shown in Figure 3b the input to the bottom-up stream is the image; the input to the top-down stream is a TD instruction, coming from the cognitive part. The instruction is composed of several vectors (typically 2-3), used to provide guidance for the next processing cycle. At the top level of the TD stream ($\overline{L^T}$ in the figure, for Top-Layer), the instruction is combined with the top-level image representation produced by the BU stream ($L^T$ in the figure), and then propagates through all layers of the TD part. This input to the TD stream is shown in more detail and explained in Figure 5 and in Supplementary S2 (The BU-TD structure), where we describe more details of the lateral cross-streams connections we use. Through the cross-stream connections, the next BU pass will now take place in the context of this TD representation. It is consequently controlled and guided by the instruction, the image context, and previous cycles. Figure 3c shows an 'unfolded' version of the structure in (b), which is used for training the network as described in the next section. We will refer to the combined network, or its unfolded version, as the BU-TD network.

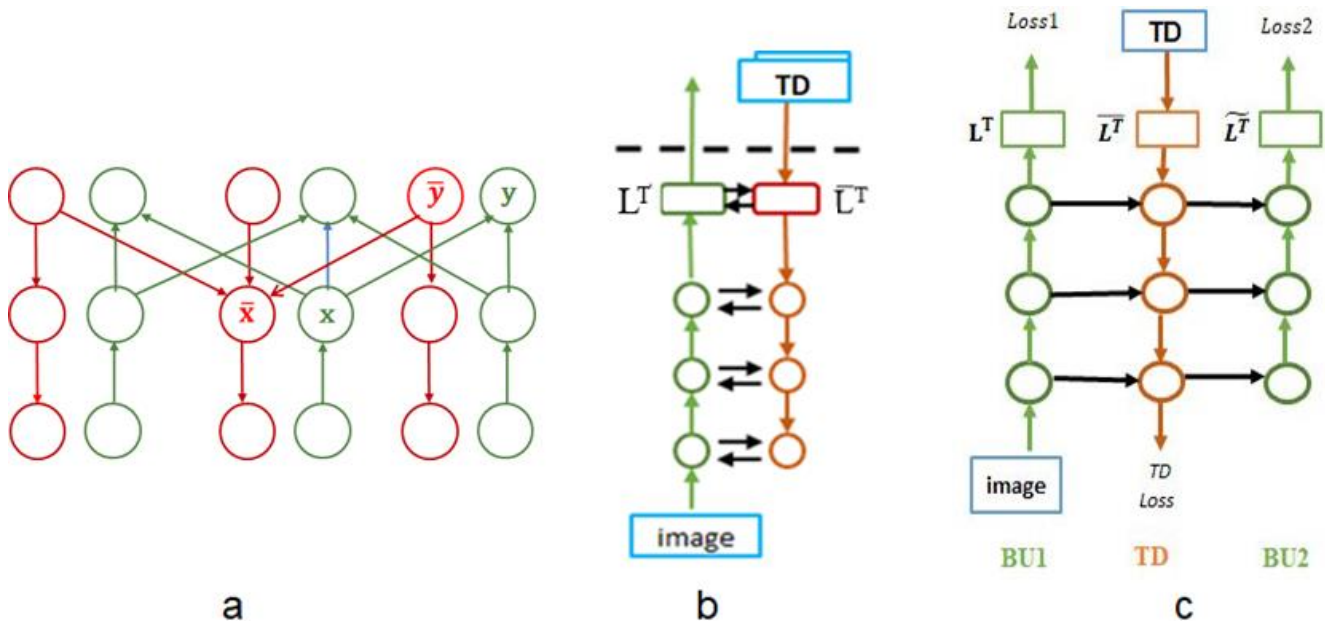

*Figure 3: Counter-streams structure* a. Interconnections of units within the bottom-up (green 'neurons') and top-down (red 'neurons') streams (cross-stream connections not shown). b. Schematic sketch of the two streams, input to the BU stream is the image; input to the TD stream is a top-down instruction from the cognitive part. The green and red circle represent layers along the BU and TD streams. $L^T$ is the top-layer of the BU stream, $\overline{L^T}$ the input layer of the TD stream. c. Training of the BU-TD network, using 'unfolding' in time. The network column BU2 is the same as BU1, at the next time step; BU1, BU2 have identical weights. The final output is at the top of the BU2 part ($\widetilde{L^T}$), and the training error is measured by Loss2. There are two optional auxiliary measures, at the end of BU1 (Loss1) and the TD stream (TD loss). See more details in Figure 5 (Full-structure).

Following the training process, instructions to perform different visual tasks can be provided as a top-down instruction input of the TD network to select and guide the next processing stage. As discussed in later sections, the BU visual stream will then perform selectively the instructed task in an efficient manner, which generalizes well to related tasks. Furthermore, sequences of TD instructions can be applied as programs to extract structured visual representations, as discussed in Section 4 (Extracting scene structures using BU-TD sequences).

Before describing the network, training and function in detail, Figure 4 (TD instructions) below shows a simple example of what it means to provide the network with a top-down instruction and obtain an appropriate response. The input images (example in Figure 4) contain a variable number of synthetic person-faces (face, neck and shoulders), from our Persons dataset, where each person (identified by a number, or a name, which the model was trained to recognize), is defined by several unchanging face features, combined with variable properties, such as different eyeglasses, shirts, hairstyles etc. (see Supplementary S3 (The Persons data set)).

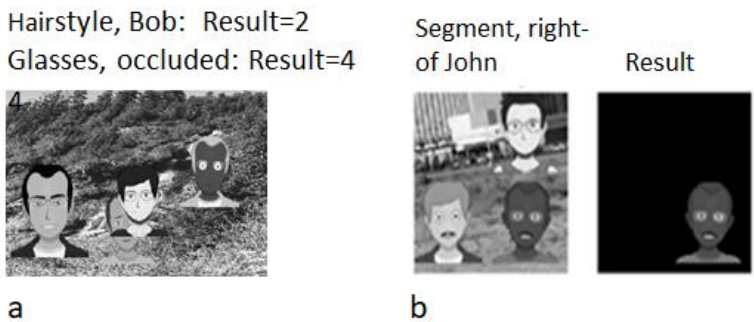

*Figure 4: TD instructions* Example of TD instructions and model responses. a. Different TD instructions with the same input image: an instruction to classify the hair-style of a particular person ('Bob') produces Hairstyle = 2, an instruction to classify the glasses-type of the occluded person produces Glasses-type = 4. b. TD instruction to segment the person right-of John produces the segmentation image on the right.

The network is trained by a procedure described below to extract a particular property of a selected person. Suppose that the current goal is to instruct the model's visual stream to extract the hairstyle of Bob, one of the persons in the image. This is obtained by supplying a TD instruction to the top-down stream, which specifies by two learned instruction vectors (i) a task to perform (the property to extract), and (ii) an argument, i.e. what to apply the task to (e.g. Bob above). As explained further below, the TD net learns to use the instruction vectors, to guide the TD-BU cycle to apply the task and argument specified by the TD instruction. In analogy with human vision (Fink *et al.*, 1997), the argument can be specified in the counter-streams using the person's identity, similar to object-based attention, or by its location, similar to spatial attention, as well as by a relation (e.g., 'right-of', 'occluded-by'). For the same input image, different tasks can be applied, by selecting the appropriate TD instruction. For the instruction 'hairstyle of Bob' (Figure 4a), the model produces the correct answer 'type 2'. Other properties can be extracted, one at a time or in combinations of several properties together, applied to different persons (see Section 3.2 (Properties)). The processing can also be directed towards occluding or occluded objects, e.g. for the instruction <Glasses type, occluded person>, glasses type '4' is identified. In Figure 4b the TD instruction was to segment the person right-of John. The output is a segmentation map, produced in this case at the bottom of the TD stream.

### 2.2 Learning in the counter-streams model

We describe next how an instruction is provided to the counter-stream net, and how the net learns to follow the instruction, and produce the instructed output. Training in the BU-TD model is accomplished using standard methods for recurrent versions of deep networks (P.J. Werbos, 1990; Lecun *et al.*, 2015). To deal with recurrent nets (RNN), backpropagation-based training is extended using 'backpropagation through time', which turns the recurrent net into a standard feed-forward model, using 'unfolding' of the network in time, shown in Figure 3c. This is an unfolded version of Figure 3b, showing three columns, where the last column is identical to the first, but at a later time. The model is given simultaneously both an input image and a TD instruction, and the training is then used to extract the output specified by the instruction. The left column marked as BU1 in Figure 3c is the first BU pass through the net, followed by TD (the top-down pass), and then BU2, which is the subsequent BU pass. Since both BU1 and BU2 use the same bottom-up circuit, their weights in the unfolded net are constrained to be the same. In the unfolded network, the output is produced at the top of the BU2 pass. In the recurrent net, where BU1 and BU2 use the same network, the output is read at the top of the bottom-up part, at the end of the BU-TD-BU cycle. The BU-TD model is a recurrent network, which can be used for multiple cycles. The unfolded version in Figure 3c trains the networks for a single BU-TD-BU cycle. However, the network can also be trained for multiple cycles, as described in supplementary S2 (The BU-TD structure, Training multi-cycles).Training proceeds using a set of training images, each combined with a TD instruction. For the images in Figure 4 (TD instructions), examples will include instructions to extract a selected property of a particular person, segment a selected person, compute a given relation between persons, and the like. The training is used to learn the network weights, including the weights in the top-down and cross-streams connections. Figure 5 (Full-structure) shows a full diagram of the unfolded network, marking the part where top-down instructions are provided to the TD stream (red frame). This part of the net, magnified in Figure 6 (Providing instruction) is described in detail in the next section. Additional details of the full structure, including the cross-stream connections, are described in Supplementary S2 (The BU-TD structure). The BU-TD network in Figure 5 is based on ResNet-18 (He *et al.*, 2016), however any standard deep network can be used for the BU part of the BU-TD structure, and the full BU-TD network can be created by adding the complementary TD network and the lateral connections illustrated schematically in Figure 3 above.

The TD instruction is provided to the network model by two input vectors, one specifying a task to perform (denoted $I_{task}$), the second (denoted $I_{arg}$) specifying an argument, or the object to apply the task to, as illustrated in Figure 6 (Providing instruction). In the examples shown in this section, the task $I_{task}$ is specified to the model during learning using a 'one-hot' vector representation, where each vector entry stands for a particular task. For instance, a single '1' in the k'th entry stands for task-k (in a list of possible tasks). A '1' at one location may specify the task of classifying the glasses-type, another location classifying shirt-type, or to perform object classification etc. As discussed further below (Section 5.3 (From symbolic to embedded instruction representation)) this can be considered a 'purely' symbolic representation form, in the sense that it designates a task without supplying information about the task or its relation to other tasks. The argument $I_{arg}$ was similarly specified using a one-hot vector, e.g. for the different persons presented to the model. Some variations are described below, in particular, when the argument is an image location, it can also be provided by a spatial map together with the input image, and in some cases, such as dealing with certain relations discussed below, the argument includes two vectors rather that one.

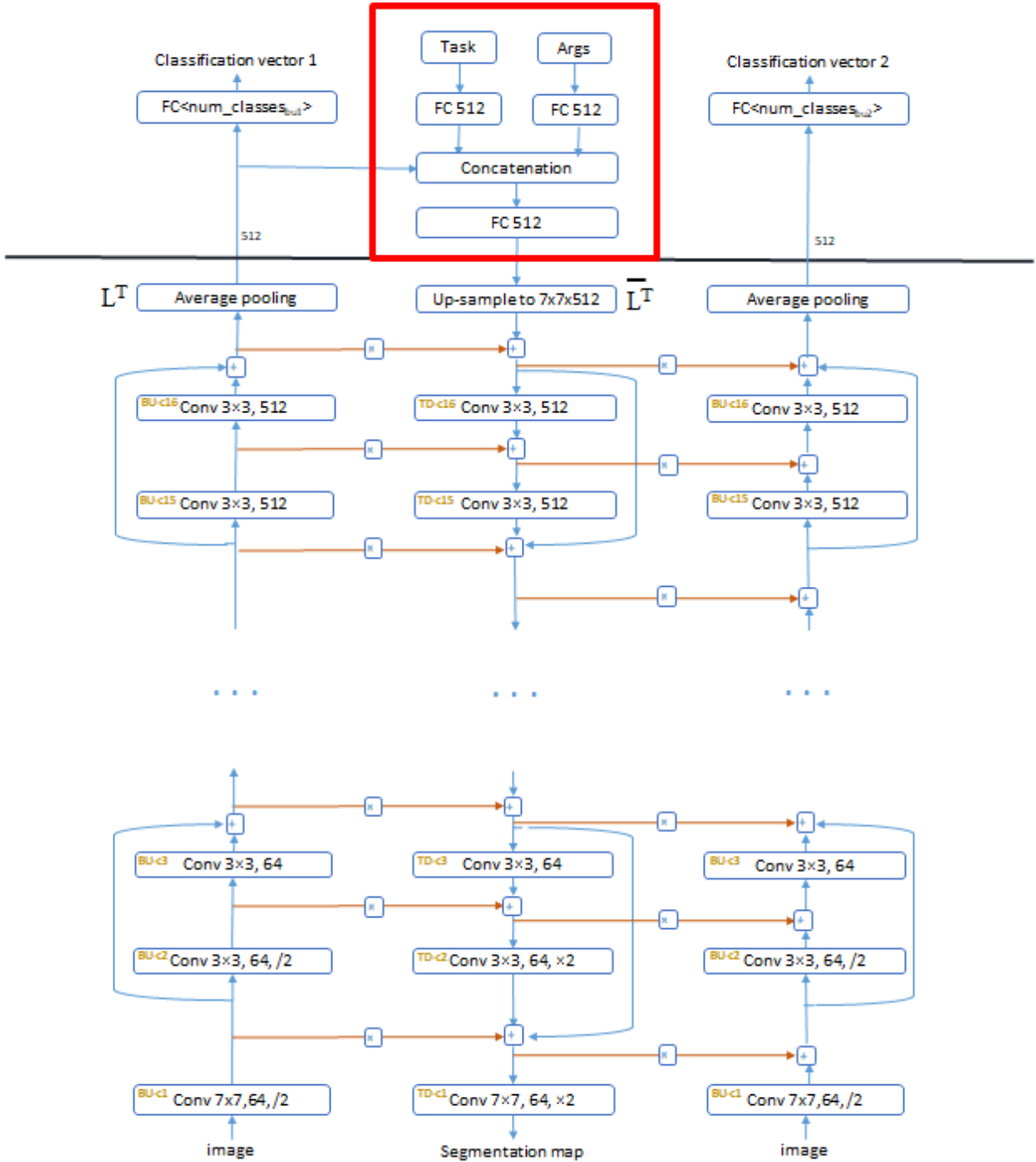

*Figure 5: Full-structure* Full structure of a BU-TD network. The BU network in this example is a ResNet-18. The red frame surrounds the part where the TD instruction is provided, as detailed in Figure 6 (Providing instruction). Each layer shows the layer identifier (such as BU-c3), the operation it performs (conv for convolution, FC for fully-connected) and dimensions; small boxes with + or x indicate additive or multiplicative operation.

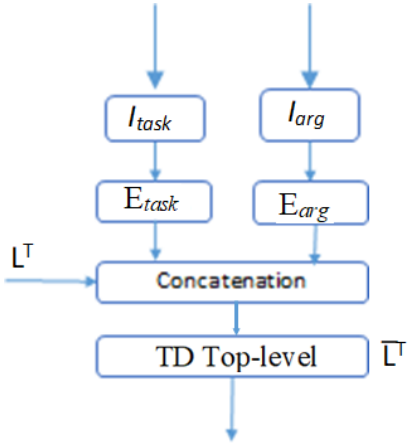

*Figure 6: Providing instruction* Providing a TD instruction to the BU-TD network. $I_{task}$ $I_{arg}$ are vectors providing the task and argument in a symbolic representation (e.g. a one-hot vector). $E_{task}$, $E_{arg}$ are embedded forms of the task and argument, produced from the previous representation by learned weights (fully connected). $\overline{L^T}$ is the top-layer of the TD stream, and $L^T$ is input coming from the top-layer of the bottom-up stream, BU1.

During training, the 'one-hot' vectors $I_{task}$, $I_{arg}$ are transformed into a new learned representation (denoted $E_{task}$ $E_{arg}$, 'E' for 'embedded'). The network weights creating this representation (which are initially random) are learned during the training period. As discussed below, unlike the initial symbolic representation, the transformed 'embedded' representation carries some useful information about the represented objects.

These two learned instruction vectors, together with the top-level representation of the BU stream ($L^T$ in Figure 3 (Counter-streams structure)), all of the same dimension, provide the input to the top-level representation of the TD stream ($\overline{L^T}$ in Figure 3). The network was trained in the examples of Figure 4 using two loss functions (using a single loss, or an additional auxiliary loss are also possible, see Supplementary S2 (The BU-TD structure). The first loss is applied at the end of the first BU stage (BU1). For extracting person-properties, the network BU1 was trained to identify all the persons (up to 6) in the image. The desired output at the next BU stage (BU2) was the correct property of the selected person, and the loss used in training was the distance (e.g. cross-entropy) between the correct answer and the one produced by the net. A third possible loss used in training is at the end of the TD stream (Figure 3), used for example for segmentation tasks.

During training, the BU-TD scheme learns to use the instruction vectors $E_{task}$, $E_{arg}$ and guide the subsequent BU pass to extract the required property of the relevant argument (e.g. Bob's eyeglasses). After the training process, instructions to perform different visual tasks can be provided to the TD network, and the visual stream will then perform selectively the instructed task, applied to the selected arguments. Additional details on the network structure and training are described in Supplementary S2 (The BU-TD structure).

Code and data are available at https://github.com/liavassif/BU-TD.

## 3. Top-down instructions

As discussed in the general introduction, to deal with complex structures in the image, the visual process is required to identify scene entities such as objects and people, together with their individual properties, as well as relations and interactions between them. We include below examples of extracting selected objects, properties and relations. The same network can be instructed to perform different tasks, and apply them to selected image entities. In subsequent sections, we show how the individual instructions are combined to form instruction sequences and forms of 'visual routines', which can extract complex scene structures, efficiently and with broad generalization.

### Top-down instructions for objects, properties and relations

We discuss below the instructed extraction of object, properties and relations, including generalizing relations via location, and the use of referring instructions.

### 3.1 Objects

In extracting and identifying objects in an image, the most frequent tasks are object classification, localization, and segmenting the object from its background. These three aspects are essential object characteristics, which are needed for almost any visual task. In human vision, perceiving an object is usually coupled automatically with identifying what it is (its basic class), its location, and separating it from the background (Rosch *et al.*, 1976; Grill-Spector and Kanwisher, 2005). In computer vision, the initial steps applied to a multi-object image, often include classification, localization and segmentation of some or all of the objects in the image (He *et al.*, 2017).

The counter-streams network can be instructed to produce classification, localization and segmentation and combine them in different ways. For example, in Figure 7a, the task is to segment a selected person. In the first bottom-up pass (BU1), prior to any TD instruction, the model was trained to produce the names of all persons present in the image (we assume that a person does not appear in the image more than once). To segment a selected person from the background, the model was trained using the instruction <segment, person>, specifying that the task to perform next is object-segmentation, and 'person' used here is the person's identity as obtained in BU1 (e.g.: <segment, person-5>). The task to perform, and the person's identity, are represented by one-hot vectors. The segmentation results are created at the bottom of the TD pass, rather than by the next BU pass. In Figure 7b the task is again segmentation, but the argument provides a location (such as top-right) rather than a selected person. In Figure 7c,d the arguments for segmentation are supplied by a property – segment the occluded person (in c), or the closest-person to the camera (in d). The image in (d) is taken from our Actors data set, which allows the creation of computer generated complex scenes with multiple persons and objects, described further below and in Supplementary S4 (The Actors data set).

The segmentations produced by the network are 'instance' segmentation, rather than so-called 'semantic' segmentation, which lumps together all objects in the image that belong to the same class, e.g., all persons in the image are considered a single image segment (Shelhamer *et al.*, 2017). It is worth noting that segmentation of different classes and at different locations are produced by the same BU-TD network, by using different TD instructions.

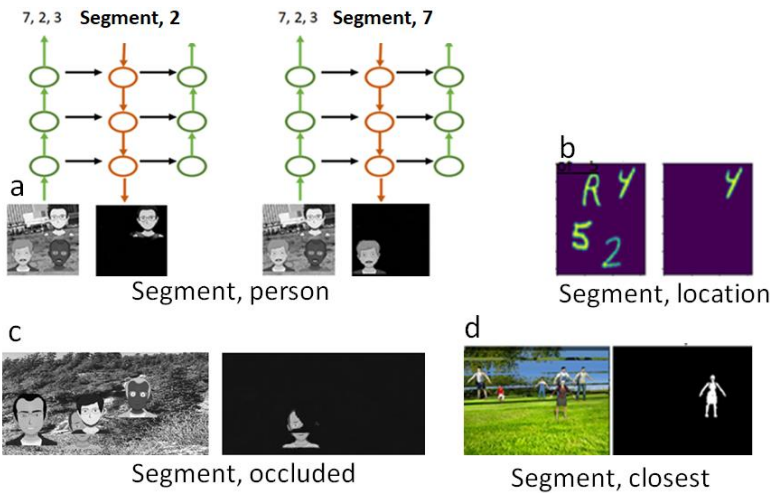

Segment, person

Segment, location

Segment, occluded

Segment, closest

*Figure 7: Selective segmentation* Selective segmentation. The TD instruction is shown under each image. a. Segmentation by a person's identity. The same image, with different arguments produces different outputs: the segmentation of the selected persons by their identity, (person-7 or person-2), at the end of the TD stream. b. Segmenting by location, e.g. segment top-right, applied to MNIST digits. c. Segmentation by property: occluded person (in c) or person closest to the camera (in d).

In the following example below, the visual stream is guided to perform object classification at a selected location. This form of selection (Treisman and Gelade, 1980), guiding the visual process to a selected location, is often referred to as 'the spotlight of attention' (Buschman and Miller, 2010). In the BU-TD model, the guiding instruction in this case is of the form <classify, location>. The location can be specified in different ways, in particular, by the object's bounding box, its segmentation mask, or its central location.

An example of classifying by location is shown in Figure 8a, applied to so-called Multi-MNIST images (Sabour, S., Frosst, N., and Hinton, 2017; Sener and Koltun, 2018), in which multiple partially overlapping MNIST images are placed on the same image. We used 2-9 digits, placed with partial overlap in fixed sub-regions of the image, and the task was to classify the digit at the location indicated by the TD instruction. The BU-TD net was based on LeNet (LeCun *et al.*, 1998), using 60,000 examples at each location. Details of the network, procedure and results are described in (Levi and Ullman, 2020). Classification accuracy for the challenging 9-digit case was 88%, which was similar to the accuracy obtained by 9 individual LeNet networks, each trained on one of the 9 locations (86.6%).

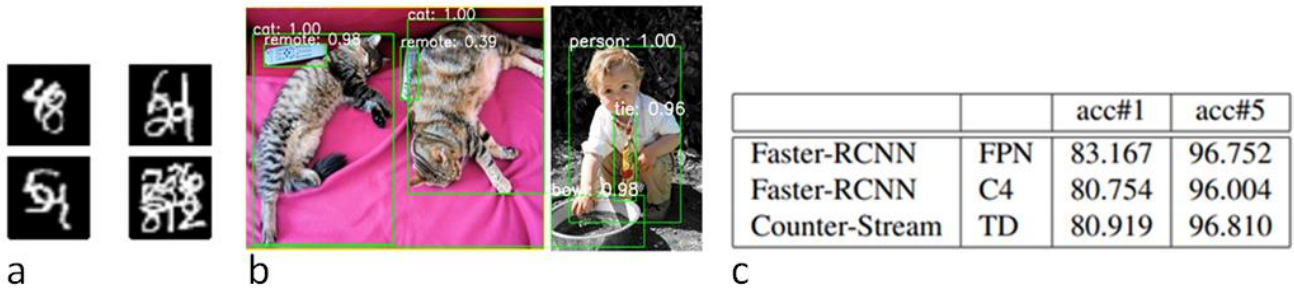

| | | acc#1 | acc#5 |
|---|---|---|---|
| Faster-RCNN | FPN | 83.167 | 96.752 |
| Faster-RCNN | C4 | 80.754 | 96.004 |
| Counter-Stream | TD | 80.919 | 96.810 |

a                          b                          c

*Figure 8: Classify-location* Classification by location. a. Classifying a digit at one of up to 9 locations. b. Classifying each object within a bounding box, given together with the input image, results with a confidence rating is shown at the top of the frame. c. classification accuracy, compared with state-of-the-art methods. acc#1 and acc#5 are accuracy of the top-1 classification and top-5 classification, respectively. FPN and C4 are two variants of the Faster-RCNN model (He *et al.*, 2017), see Supplementary S5 (Classification by location) for details.

Figure 8b shows results of classification by location on natural images. For each object, the correct location was supplied to the network together with a classification instruction. The locations in (b) were provided in a TD manner, by a coarse map of the object's bounding box map (shown by the green outline). The classification results with a confidence rating are shown at the top of each frame. We compared several alternative methods of supplying the location information: location of the object's center (x,y coordinates), coordinates of the bound box, the bounding box as a spatial map, or the object's segmentation map. Of these, the bounding box map produced the best results. We further tested the possibility of providing an object's location in a BU manner, together with the input image. We found that providing a bounding-box map, or a segmentation map, can also be used with similar accuracy. The classification results, shown in (c), are close to the best reported results. Further details of the implementation in comparisons are given in Supplementary S5 (Classification by location).

### 3.2 Properties

The next examples illustrate the guided extraction of object properties. Figure 9a shows the extraction of selected properties associated with different people in the image. Images were taken from the Actors data set (Section 4.1 (Extracting scene descriptions), Supplementary S4 (The actors data set)). Given the same image, different TD instructions will guide the model to focus on different properties of the same or different persons, such as wearing glasses, hair-style and others (we use here 'wearing glasses' as a property). The person to extract the property for was given in this case by its segmentation map, extracted by the model, and given as an auxiliary input at the bottom of BU2 (described in more detail in Section 4.1 and Supplementary S7 (Extracting full scene structure)). The process of property extraction is then repeated to extract additional properties of the same person, and then additional persons are selected, and their properties are extracted, as shown in the figure. More details and results related to the extraction of properties and relations in the Actors data set are described further in Section 4 (Extracting scene structures using BU-TD sequences).

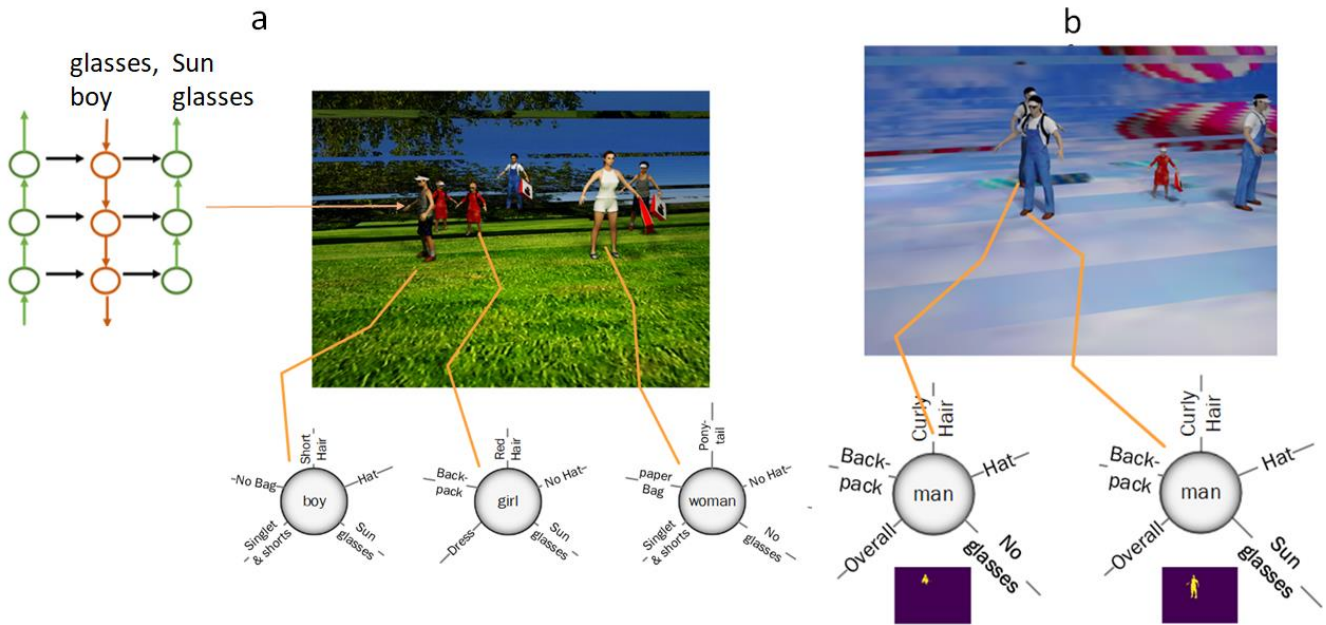

*Figure 9: Extracting properties* Extracting selected properties from selected persons: a (left). The extraction of a single property, by a TD-instruction to extract the 'wearing-glasses' property of a selected person, the instruction to the network is shown with arrow pointing to the boy. a (right). The process of property extraction is repeated to extract all properties of the same person, and then additional persons are selected, and their properties are extracted. b. Extracting properties from two highly overlapping persons. Extracted segmentation maps of the persons is shown at the bottom.

*Binding properties and objects*: As illustrated by the figure, by selecting objects (or people) and their properties, it becomes straightforward in the model to make the correct association between objects in the images and their properties. This association is a part of so-called 'binding' in perception (Malsburg, 1995; Feldman, 2013), discussed further in Section 5.2 (Combinatorial Generalization). Figure 9b shows the extraction of properties from two highly overlapping persons. The guided extraction of properties from different objects does not require spatial separation between them, and properties of different objects within the same region of space are correctly associated with the corresponding objects.

*Sequential and parallel extraction*: In the example above, the model was trained to extract a single property at a time, and property-extraction instructions were then repeated to extract multiple properties of different objects, e.g. the selected persons in Figure 9a. The selection of properties to extract need not be limited to a single property; instead, multiple properties associated with an object can be extracted simultaneously using a single instruction. A representation identical to Figure 9 of the persons and their properties was also obtained by training the network with an instruction to extract all the properties of a selected person. In extracting multiple properties of the same object, it is also possible to group together related properties, such as 'clothing' or 'pose' properties of a person, and extract each set using a single instruction.

*Compound instructions:* Finally, we also found that training to extract individual properties of an object can generalize spontaneously to the extraction of more than a single property at a time, without additional training (Wagner, 2020).

As described above, the instruction to extract a specific property used a one-hot encoding by a property-vector: the selected property is represented by a single '1' turned on in the representation. Following training, using an instruction with a '1' turned on at two vector locations, never used together in training, leads to the extraction of the two selected properties simultaneously. For example, combining the instructions for extracting glasses-type and hair-style into a single vector will lead to the correct extraction of both properties of the selected person. Compound instructions can be also extended to a larger number of properties with high accuracy. We tested compound instructions on our Persons data set, training for individual properties only, and then testing with compound instructions. In training and extracting each property individually, average accuracy was 96%. Extracting all properties simultaneously, without additional training yielded 94% accuracy.

We found that such compound instructions can be used for extracting simultaneously two or more different properties of the same person, but not for extracting simultaneously the same property for two or more different persons. This asymmetry (properties can be combined, but not the persons), is not surprising. In extracting, e.g., glasses-type of two persons, there is an additional problem of combining the persons with their correct properties, a problem that does not arise in extracting two different properties of the same person. We further found that, similar to the extraction of a single property, extracting multiple properties by compound instructions generalizes well to new object-property pairs (discussed in more detail in Section 5.2 (Combinatorial Generalization)). Finally, by performing ablation studies of the network, we found that the generalization to compound TD instructions depends on the cross-stream lateral connections. Repeating the same experiment (training on single instructions and then applying compound instructions to readout all the properties together) without the cross-stream connections reduces accuracy from 96% to 84%. Limiting the ablation to lateral connections in one of the two directions, we found that BU to the TD stream played a larger role than the connections from the TD to the BU stream.

### 3.3 Relations

Relations of interest in a scene include interactions between people, between people and objects, and between objects. An example of persons' interactions shown in Figure 10a is a 'facing' relation, between two people facing each other, typically at a close proximity, taken from our Actors data set. We trained for the instructed extraction of 'facing' using an instruction of the form <facing, person-1>, where person-1 is a selected person in the image. A network for the Actors set was trained to both segment and classify the person who is facing person-1 (or to indicate that there is no one facing person-1). As discussed above (in discussing classification and segmentation), an object can be specified in the TD instruction either as a TD vector, or using an auxiliary input image. In the Actors data set, the reference person (person-1 above) was supplied to the network using an auxiliary segmentation map, given as an additional channel in the input image. The output was a segmentation map produced at the end of the TD pass, and a classification produced at the end of the subsequent BU pass (Supplementary S4 (The Actors data set)).

Figure 10b is an example of the same network trained for extracting a 'touching' relation between people in the image. Additional relations and how they are used during scene analysis is described in more detail in Section 4.1 (Extracting scene descriptions).

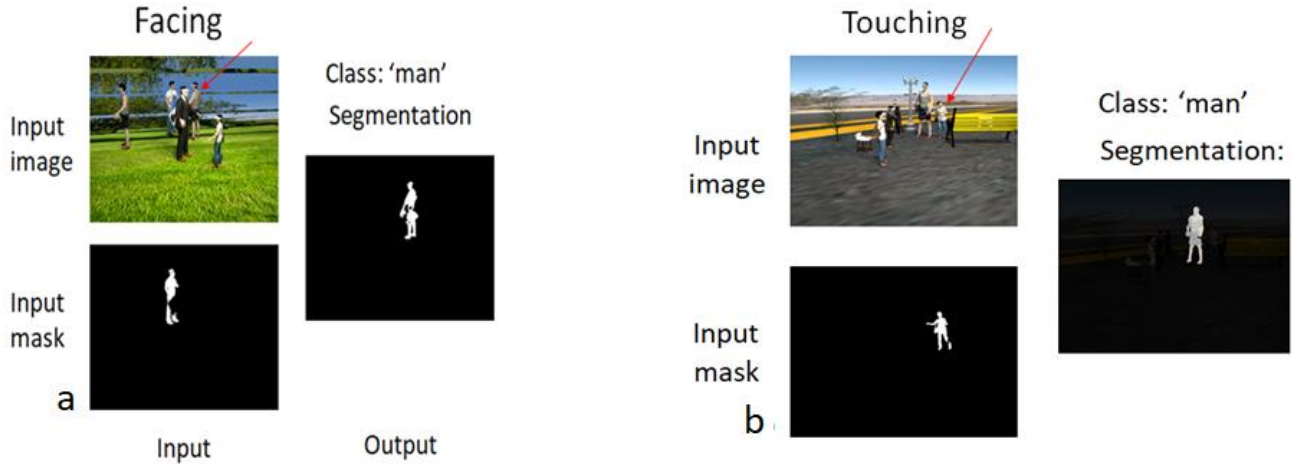

*Figure 10: Interacting people* a. Extracting the relation 'facing', using an instruction <facing, person-1>. The red arrow points to the facing people at the back. The reference person (person-1) is given by a segmentation map (input mask) together with the input image, and the output is a classification and segmentation of the target person. b. The relation 'touching'. Red arrow points to the interacting people.

The extraction of relations in these and other examples can also be trained on an alternative form of providing the input and computing the relation. Instead of providing one of the participants in the interaction and locating the second, the alternative, and more standard mode, is to train the model with two persons, x and y, and a relation R (using two instruction arguments), and determine whether or not R(x, y) holds. An example is shown in Figure 11 (Facing(x,y)) below. The network was trained on instructions of the form Facing(person-1, person-2). Two input persons were provided using segmentation maps, provided with the input image. The output is a binary choice 'facing' or 'not-facing'. Another example, of the relation Occlude(x,y) is illustrated in supplementary S6 (Spatial relations). The use of the two forms of using TD relation instructions is discussed further below, in the subsection Referring instructions.

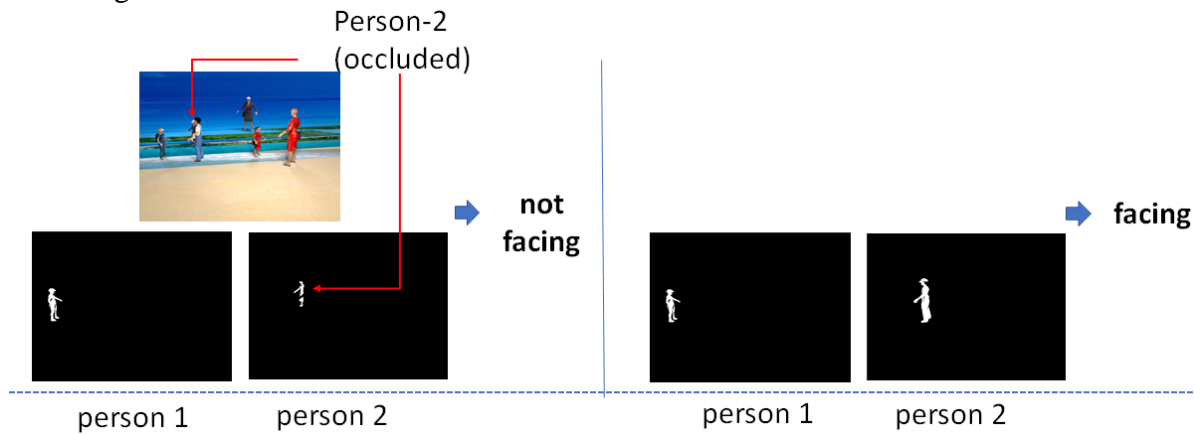

*Figure 11: Facing(x,y)*  The model computes the relation Facing(person-1, person-2) between two persons in the image. The input image is given together with two segmentation maps of the two persons (bottom row). For the selected persons on the left, the output is 'not facing; note that person 2 (pointed at by the red arrows) is highly occluded. For the persons selected on the right, the relation 'facing' holds.

## Human object interactions

Examples of human-object interactions are shown in Figure 12 (Holding) for the relation 'hold'. The model classifies the held object (such as hammer, hand-held sign, bag, racket), and produces its segmentation map.

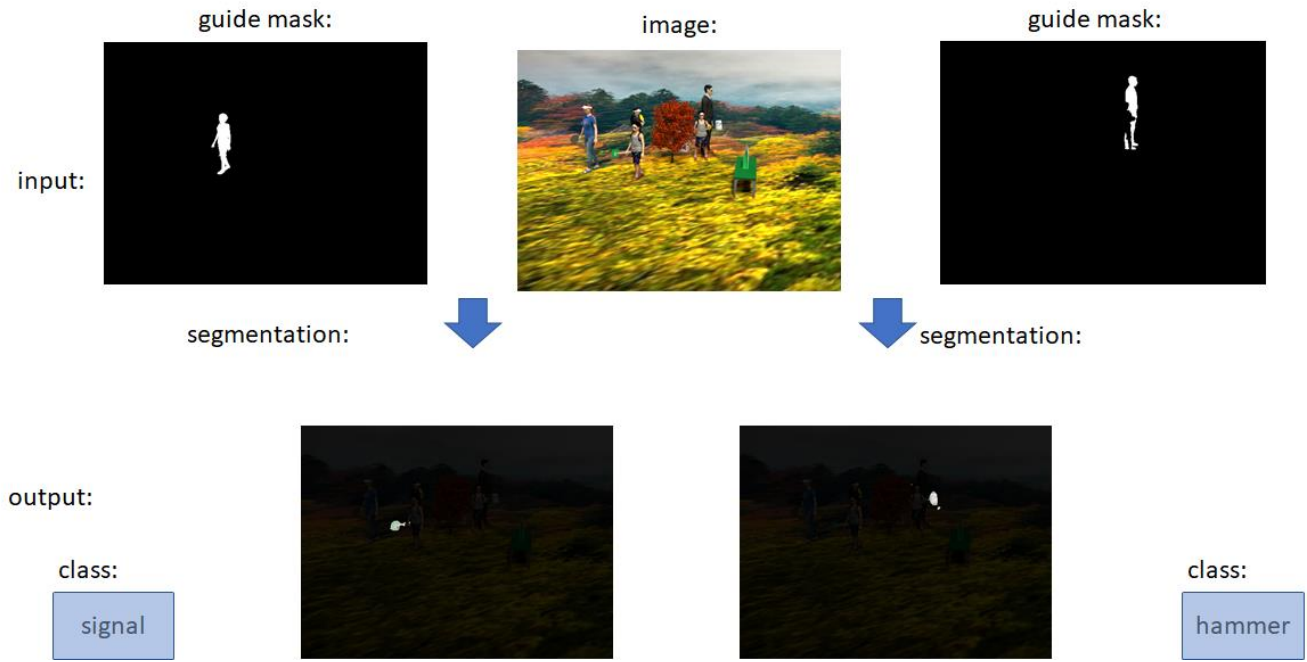

*Figure 12: Holding*  Extracting two 'holding' relations from the same scene. Top row shows the input, the scene image and the guiding masks of the reference persons for the relation. Bottom row shows the segmentation maps of the held objects

## Object relations

Figure 13a shows the extraction of the relation 'part-of' between objects and their parts. Images were taken from the CUB200 data set (Wah *et al.*, 2011), which is a fine-grained recognition dataset that provides bird images of 200 bird species, with 15 annotated bird parts such as head, eyes, crown, throat, leg, and others, together with a set of attributes. Parts are annotated in the data set by a center point for each visible part.

The BU-TD net was trained in this case to produce the part location (a heat map at the bottom of the TD stream), and the color of the part (at the output of the BU2 pass). The different parts were treated as different tasks, each specified by its TD instruction one-hot vector. Accuracy of the BU-TD model in predicting the correct property was 80.9% compared with 74.3% obtained by training an individual network for each task (more details in (Levi and Ullman, 2020)).

The ability to select parts of interest is important for dealing with complex scenes. In some approaches to scene processing, the initial stage includes the identification (classification and segmentation) of all recognizable scene components of potential interest (Redmon *et al.*, 2016; He *et al.*, 2017; Yang *et al.*, 2018). However, the large number of parts and sub-parts of all the objects in a complex scene makes such an approach infeasible, and therefore the ability to select parts at multiple levels becomes essential.

Figure 13b below shows the extraction of 'On' relation between two objects. The scene contains an object placed on another, the 'reference' object. A TD instruction is given as an 'On' relation with a reference object map, and the result is the class and segmentation map of the top object. Other objects relations include spatial relation, which apply to both people and objects, with examples given in the next section.

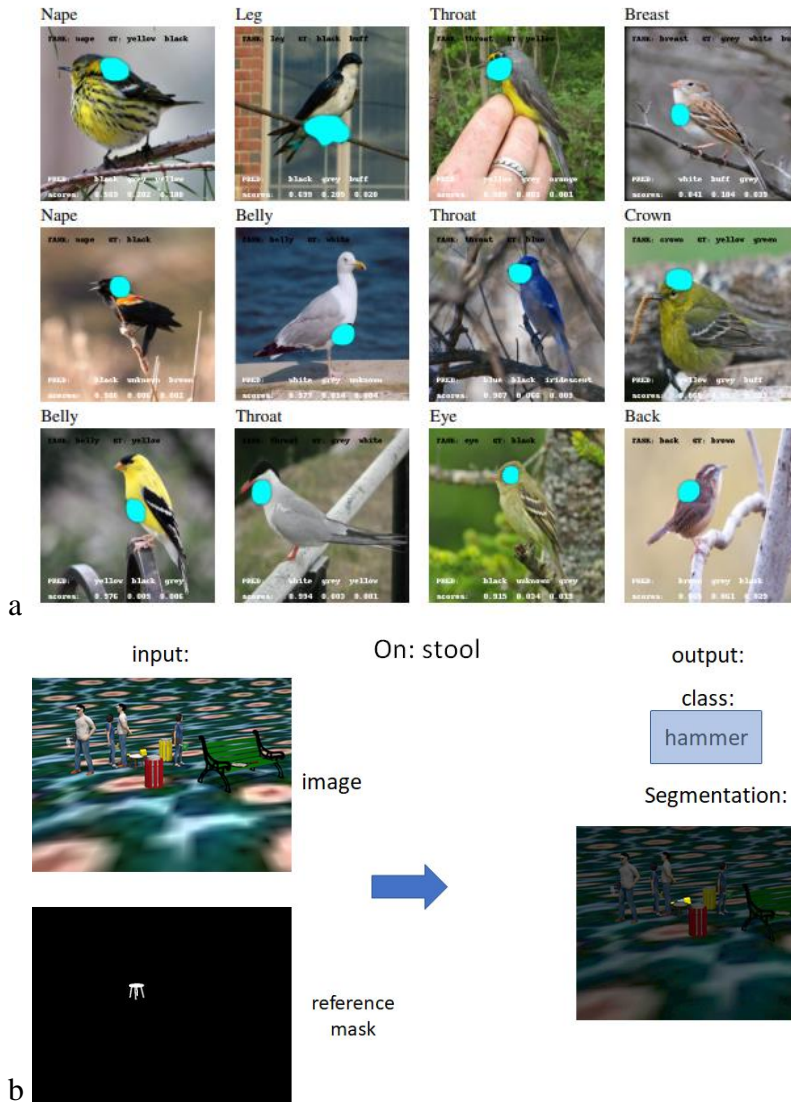

*Figure 13: Object relations* a. Object-part relations: locating parts and extracting their properties. Images come from CUB200 with 200 bird species and 15 annotated parts. A single network is used, and each part is detected using its TD instruction. Prediction of the part location is produced at the bottom of the TD stream. b. The relation 'on' between objects. Images come from the Actors data set, between objects contained in the scene, such as a tool placed on a table, bench, stool or chair. Details of extracting relations in the Actors dataset are described in Section 4.1 (Extracting scene descriptions) and Supplementary S4 (The Actors data set).

*Spatial relations*
Figure 14 (Spatial-2D) shows the extraction of spatial relations such as right-of, left-of, above and below. It shows examples from images coming from the EMNIST data set (Cohen *et al.*, 2017), CLEVR set of solid objects (Johnson *et al.*, 2017), and the Actors set. The first BU pass (BU1) was trained on classifying the objects in the scene (i.e., which object-classes are present), and the TD instruction was to either classify or segment a selected object in the image. In most cases, the object in the instruction was specified by its class (assuming that a single object of this class was present in the image), but other methods of specifying the reference object were used as well. If there is no neighbor in the scene satisfying the relation, a 'no-neighbor' is returned. Relations in the instruction were represented in a symbolic, one-hot form; similarly, the class of the reference object (the argument) was represented by a one-hot vector. The networks for the different domains were separate, and each network was trained on multiple tasks for the different relations and objects, such as <above, w> of <right-of, cylinder>. Additional details are in Supplementary S6 (Spatial Relations).

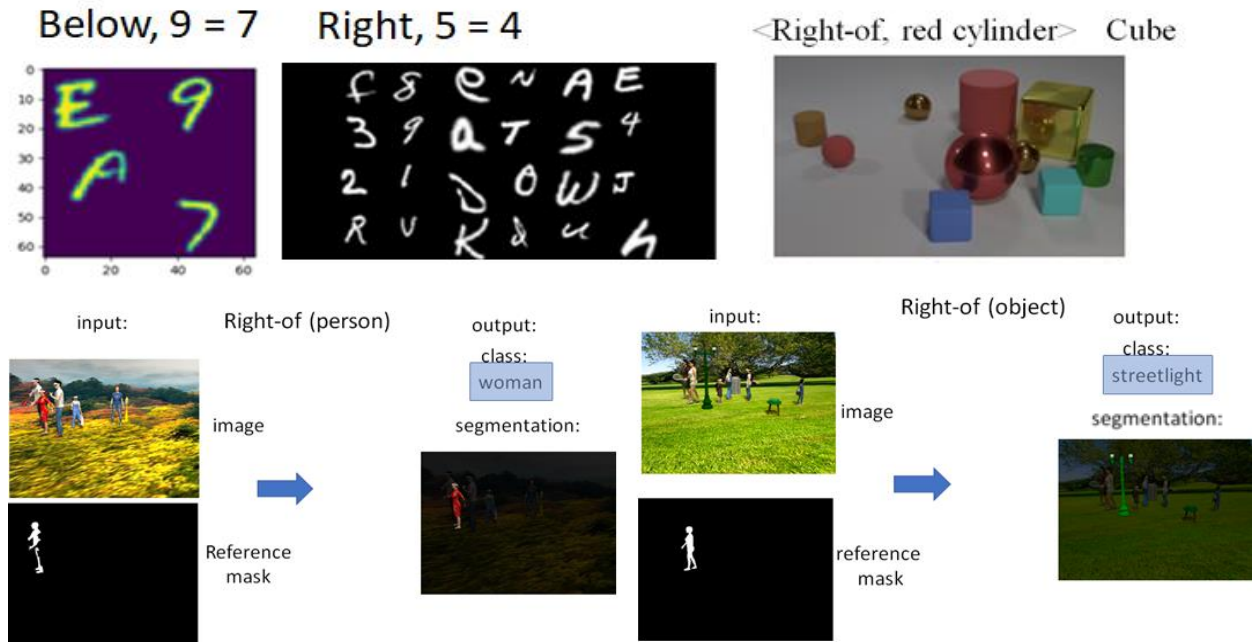

*Figure 14: Spatial-2D* Learning the relations, right, left, above, below. Examples coming from EMNIST images (top, left and center), CLEVR (top right), and the Actors dataset (bottom row). The Actors examples show right-of when the target is a person (left) and when the target is an object (right).

Figure 15 (Spatial-depth) below includes spatial relations in depth, in which an object can be occluded by another, or further-away than another object. In Figure 15a the BU-TD model was trained to segment and classify either the occluding or occluded person. We assumed at most one occlusion in the image, but the procedure can be extended to multiple occlusions. In Figure 15b the model was trained on the Actors data set to segment and classify people in the scene based on depth-order. The input to the task was a scene together with an auxiliary segmentation map. Given the segmented region of a reference person ('person-1') in the image, the instruction <behind, object> classifies and segments the next person in the image in terms of their distance from the camera. Additional details are described in Supplementary S6 (Spatial relations), and Supplementary S4 (The Actors data set).

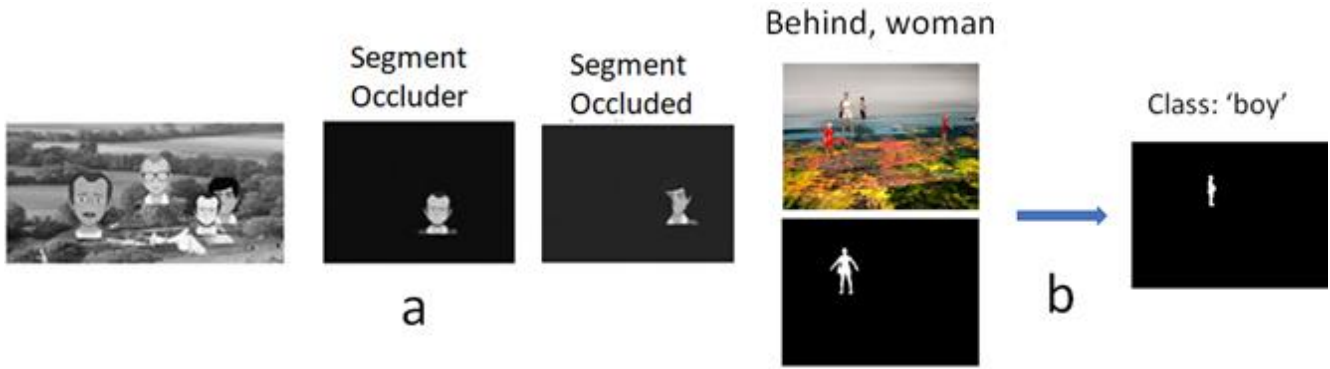

*Figure 15: Spatial-depth* a. The task is to segment a person, applied to either the occluding or occluded person. b. The relation of behind, in the sense of further-away than another person. Given a reference person by its segmentation map, the instruction <behind, object> segments and classifies the person in the image who is the next in terms of distance from the camera with respect to person-1.

### *Extracting higher-order configurations*

Using the basic instructions for left, right, above, below, the BU-TD scheme can be used to extract higher-order spatial configurations without any additional training, by applying appropriate sequences of the basic instructions. For example, we can look in Figure 16 (EMNIST-structure) for the target configuration on the right (labeled 'target'), which is composed of the character 'W' with a '2' anywhere to its left, and a '5' directly above it. In this test of configurations, we used images with multiple EMNIST characters, where each character appeared at most once in the images.

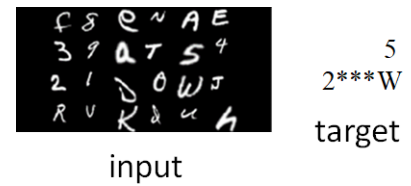

*Figure 16: EMNIST-structure*

The task is to find the target structure (a W with 5 directly above and 2 to the left) in the character-array on the left.

In our EMNIST data set, the first BU pass was trained to produce all the characters present in the image. We can next apply a sequence of TD instructions, starting e.g. with <above, w>. If the answer is '5', we apply next a sequence of 'Left-of' instructions, until either a '2' is reached (hence the target structure was found), or a row-end is reached, implying that the structure is not present in the image. This is a simple example of a general topic, discussed further in Section 4 (Extracting scene structures using BU-TD sequences), on the extraction of structural descriptions using programs composed of TD instructions.

### *Generalization via location*

Learning to extract spatial relations reveals a basic limitation of deep networks training, in terms of generalizing the relations to untrained objects. Training relations using a particular set of objects does not generalize to novel objects: for example, training spatial relations on the digits 0-9 in the EMNIST data set, fails to generalize to the characters A-Z. In contrast, for humans, a relation such as right-of is an abstract relation, which does not depend on specific object classes. We found that using an appropriate short sequence of TD instructions instead of a single instruction can lead to a broad generalization in using these relations. The basic idea behind the generalization is that spatial relations are relations between object locations, rather than the objects themselves. Instead of using a direct instruction, the relation 'right-of', for instance, was trained using a sequence of three instructions in the following manner. To extract right-of (n) for a digit n in the image, first, extract the location $q_1$ (the x, y coordinates) of the digit. Second, find the predicted location $q_2 = (x',y')$ of n's right neighbor. Third, classify the character at location $q_2$. The image inputs in tasks 1 and 3 were of digits and letters, but in task 2 in the sequence, training was limited to digit images, in order to test generalization to letters. As a further test, in addition to EMNIST characters, the model was trained on recognizing multiple additional figures (animals, symbols and Omniglot characters (Lake *et al.*, 2011), but the additional objects were never used during the training of spatial relations. The rational for this training is that tasks 1 and 3, extracting the location of an object, or classifying an object at a given locations, are general tasks that are unrelated to spatial relations and were therefore trained on the larger set. The spatial relations in step 2, however, were trained on a limited set of images.

To test generalization, the test images could include characters, or shapes from the extended set (as in Figure 17 (Relation generalization)). The results showed that the accuracy for novel objects was similar to the accuracy of the trained classes. In task 2, tested on a novel target shape, accuracy remained high (97%). The overall accuracy, using the 3 steps sequentially, was 88%, for both familiar and novel shapes (Netanyahu, 2018). For a related approach, using neuron module networks, but less general, see (Bahdanau *et al.*, 2019)

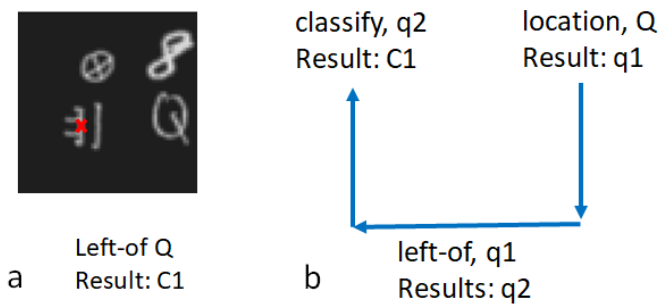

*Figure 17: Relation generalization* a. The task is to compute the relation <left-of, Q>, the answer is the shape to the left of the Q, denoted $C_1$ at the bottom-left (marked by a red x). b. The task is accomplished by a sequence of 3 top-down instruction. The first, <location, Q> produces the location (image coordinates) $q_1 = (x_1, y_1)$ of the character Q. Second, <left-of, $q_1$> produces the predicted location $q_2 = (x_2, y_2)$ of the object to the left of $q_1$. Finally, <classify-location, $q_2$> classifies the object at the specified location $q_2$. The sequence training generalizes to novel shapes such as $C_1$. The spatial relation left-of is learned between locations, rather than shapes.

Similar generalization via location could be applied to other relations beyond simple spatial relations. For example, humans can generalize the 'riding-animal' relation between a person and an animal, from known to novel classes (Figure 18 (Riding animals)). Applying the approach above, the relation will again be established through an intermediate step of identifying the location of the target animal, and then classifying it, leading to the classification of animals not seen before in the riding relation. Preliminary testing ((Netanyahu, 2018; Levi, 2021), see also (Krishna *et al.*, 2018)) support the feasibility of this mode of generalization.

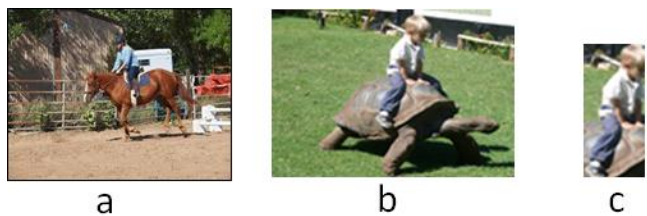

*Figure 18: Riding animals* Humans can generalize 'riding' from familiar examples, such as in (a), to novel examples, as in b. c. The relation itself can often be perceived without necessarily recognizing the target object.

## *Referring instructions*

A property worth noting regarding the extraction of relations is that the instruction to extract a relation R can come in two forms. The first, which is the commonly used form in computing relations (e.g. (Lu *et al.*, 2016; Santoro *et al.*, 2017; Yang *et al.*, 2018)), is to specify two objects (x, y), and train the model to determine whether or not the relation R(x,y) holds. In the second, which we denote as R(x, ?y), only a single object x is specified, and the relation task is to identify an object y in the image that satisfies the required relation with x, if such an object exists. A number of examples of inferring instructions were discussed in previous sections. For instance, extracting Right-of(x, ?y), where the task is to find the right-neighbor of an object x. Similarly, Occlude($p_1$, ?$p_2$), finding the person occluded by $p_1$, Facing($p_1$, ?$p_2$), locating  a person $p_2$ facing $p_1$ etc.

The relation in referring expressions is used to identify a specific object in the image that satisfies the relation R with a reference object x. This use of relations is similar to the notion of 'referring expressions' in natural language, which are noun phrases whose function in discourse is to identify some individual object. This notion is also related to the concept of 'predictive join' (or pjoin) proposed by (Papadimitriou and Vempala, 2015), extending (Valiant, 1994, 2005), as a general basic operation in cortical computation. Referring expression were also studied in combinations of language with vision. The goal is either to identify an object in the image based on a referring expression in natural language, or to generate a language expression that will uniquely identify a given object in the image (Mitchell *et al.*, 2013; Mao *et al.*, 2016; Luo and Shakhnarovich, 2017; Liu *et al.*, 2019). (Krishna *et al.*, 2018) used the term 'referring relationships' in vision, as a specific type of referring expression, where the goal is to identify a specific object in the image based on a given relation with another entity. For instance, 'the boy wearing a hat' can distinguish a particular boy from other boys present in the same image. Our use is similar, but independent of language. We train the BU-TD network on what may be termed a 'referring instruction', which locates a target object based on a relation and a reference object supplied by the TD instruction.

As discussed further in Section 4, referring instructions, which use a relation R and an object x to locate an object y that satisfies R(x,y), are useful in extracting structural descriptions of scenes by the BU-TD model. Briefly, the reason is that in the process of scene interpretation, it is often useful to move from an object x to a novel object y, which satisfies a particular relation with x. In locating y based on x, it is often possible to tell whether an object x participates in a relation R(x,y), based on the object x, without identifying y. For example, in Figure 18c above, the partial image centered on the boy is sufficient to suggest the relation 'riding'. Similarly, we found that in human scene interpretation (Supplementary S1 (The psychophysical timeline study)), an action is often suggested based on the actor alone, and the object of the action is identified only later. For example, In Figure 19a, 'blond woman' and 'drinking' are recognized first, and 'glass' is recognized only 300 msecs later. In Figure 19b 'girl', 'holding' and 'pouring' are identified in this order, with the 'funnel' and 'bottle' only significantly later (more details in Supplementary S1).  This progression can be used in scene interpretation, by suggesting the next TD instruction to apply e.g. in action recognition, where the likely action can be recognized first, followed by a TD instruction to find the action-object.

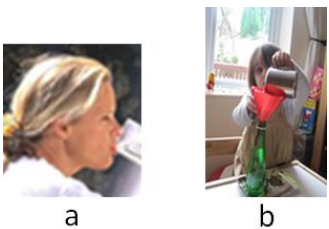

a                 b

*Figure 19: Referring-actions* In images of a person performing an action using an object, the person and action are often recognized well before the object. a. 'blond woman' and 'drinking' are recognized first, and 'glass' only 300 msecs later. b. 'girl', 'holding' and 'pouring' are identified in this order, 'funnel' and 'bottle' only significantly later (Supplementary S1 (The psychophysical timeline study)).

In extracting structures of interest from the scene (Section 4 (Extracting scene structures using BU-TD sequences)), we found that referring relations play a major role. However, the model also uses non-referring relations discussed above, of the form R(x,y), where the two objects x,y are given as arguments of the instruction, and the model is trained to determine whether or not R(x, y) holds (see Supplementary S6 (Spatial Relations)).

# 4. Extracting scene structures using BU-TD sequences

The BU-TD model makes it possible to compose complex visual tasks from sequences of steps, invoked by TD instructions. The composition of visual processes from sequences of instructions needs to be flexible in terms of tasks and images, that is, we need (i) to compose a broad range of novel tasks from the available TD instructions, and (ii) to apply a given task to a broad set of images. The selected TD instruction to apply at each step will depend in general on the goal, the image, and on information already extracted from the image in previous steps. For example, in a visual search for a particular object in a given image, the steps to perform will depend on the number of objects and their arrangement in the image. Consequently, the process of applying a composite visual task by a multi-cycle process can be considered as applying a visual program (or a 'visual routine', (Ullman, 1984)) to the image. As described briefly in the introduction (background and goals), the selection of the next TD instruction to apply may also depend on non-visual knowledge, but this aspect remains outside the scope of the current discussion. The main goal of applying visual routines to the image is to extract a relevant and informative description of the scene. In the next sections we illustrate the extraction of structural descriptions of interest using routines composed of TD instructions.

## 4.1 Extracting scene descriptions
### Full structure extraction
Figure 20 (Full scene structure) shows an example of extracting the full structure of a scene of interacting people. 'Full structure' means extracting all the scene components, their properties and relations. Since the scene was generated from the Actors dataset, the set of all components, properties, and relations visible in the image is fully known, and therefore the extracted scene description can be compared against the known ground truth.

The process of extracting the full structure is described in Supplementary S7 (Extracting full scene structure). Briefly, it proceeds along the following stages. The first stage identifies all the scene components (persons and scene objects) in the image. (Scene-objects do not include objects held by persons or placed on scene objects, which are treated separately.) This is obtained by detecting and segmenting them sequentially using the TD instruction <extract-next>. This instruction is used to both segment and classify the next person or object in the image, in order of distance from the camera. The input to this instruction includes a cumulative segmentation map (provided together with the input image), which includes all the components extracted so far. The cumulative map is initially empty, and therefore the first application of the instruction identifies the closest person or object to the camera. For each identified component, all the properties are extracted and added to the final scene representation (see below). The next component is then identified, until all persons and objects with their properties have been extracted. The next stage extracts all the referring relations in the image, for all persons-relations person-object and objects relations. Finally, the last stage can add relations by applying non-referring relations R(x,y) (Section 3.3 (Relations)) to all relevant pairs of extracted objects. For more details on the extraction of full structures, see Supplementary S7 (Extracting full scene structure).

The extraction of scene structures by the BU-TD process does not depend on training with particular structures. For extracting arbitrary scenes generated by the data set, our model required training with each of the individual instructions, namely, for the visual recognition of all the object classes, properties and relations used by the model. It also requires some additional instructions, such as identifying a new image component (e.g. to identify the next person), for extending the current structure. In natural images, the set of visual classifiers will be very large, but dealing with this large set is an inherent requirement of any scene recognition scheme. With this set of instructions, the scheme is not limited to trained scene structures, but can recover the structure of any scene. The extraction of the full structure of simple scenes is useful for testing and evaluation purposes, but for real-world natural scenes, extracting a full scene description is often infeasible and usually unnecessary, as discussed in the next section.

The structure describing the scene extracted by the process above can be considered as a graph, similar to scene-graphs (Santoro *et al.*, 2017; Xu *et al.*, 2017; Yang *et al.*, 2018). The graph structure can have the general form as described in (Battaglia *et al.*, 2018), which includes not only object properties and relations, but also global scene properties, and relations that can have attributes associated with them (such as a 'warm hug'). A general way of representing graph structures in computational models is by a collection of local descriptions, usually triplets, describing the nodes and edges of the structure. For examples, property triplets have the form: (item, property, value), (where 'item' in our implementation can be an object or a person). Relations have the form: (item-1, relation, item-2), where item-1 or item-2 are identifying numbers ('i.d.'s), generated during the extraction process.

In addition to the scene structure, the scene representation includes a 'grounding' of the extracted scene components (persons and objects) in the image. This grounding means the ability to refer back from a component in the scene description, identified by its i.d., back to the image region containing the component in question. The grounding is provided in the scene description by pairs of the form (component-id, segmentation-map), which associated each component with its segmentation map extracted from the image during the scene interpretation process. Similar to the computational scheme, the human perceptual system also faces the problem of creating some form of a structural representation of the external scene. An interesting and still open question is the form and neural representation of such structures in the human brain in general and scene descriptions in particular (Smolensky, 1990; Thomas McCoy *et al.*, 2019).

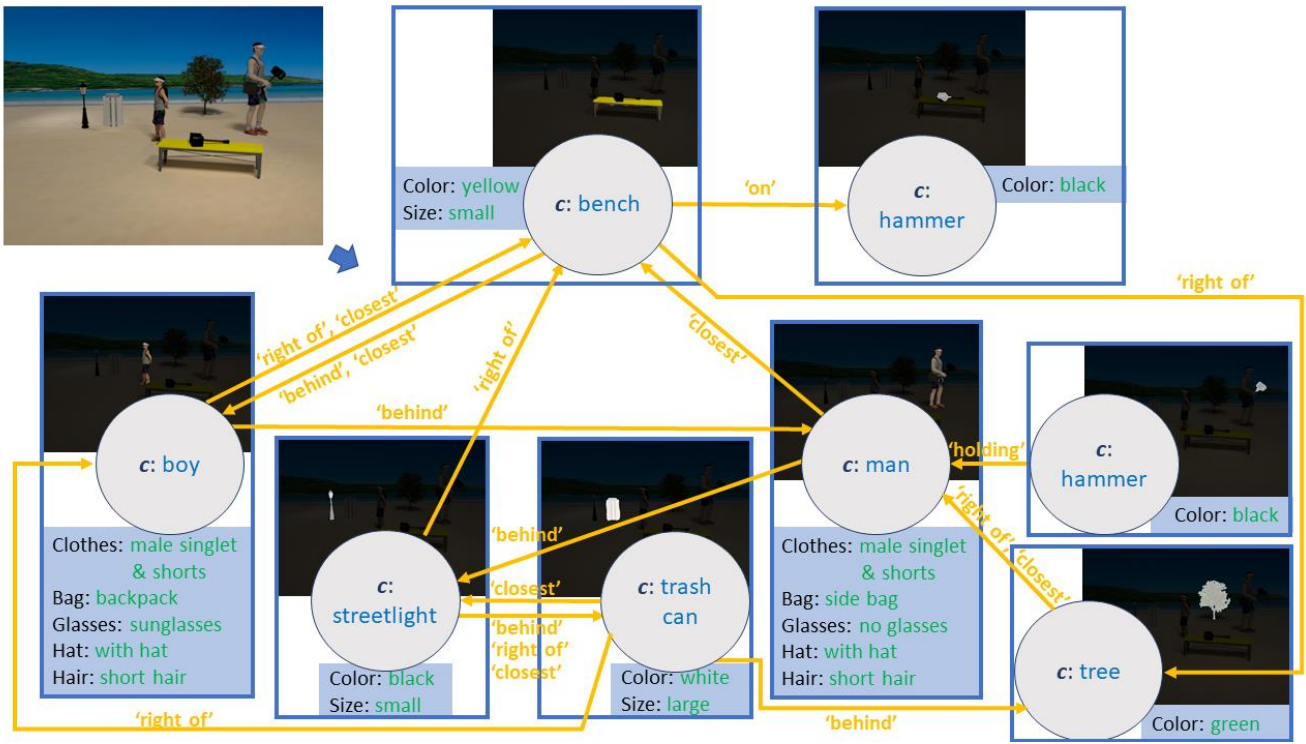

*Figure 20: Full scene structure.* The input image (top left) includes persons, scene objects and held objects. The extracted structure includes the scene component, their segmentation maps, properties, and relations between scene components. The structure was extracted by the sequential application of TD-instructions, selected automatically by the model.

### *Guided structure extraction*

In natural scenes it will be infeasible in general to extract a full 'scene graph', which contains all the scene components of potential interest, with all their properties and relations. For instance, in a scene of a room with multiple people and many objects, we may be interested in a particular aspect, for instance whether the glass held by a particular individual is empty or full. This is, of course, only one of many possibilities, and it is unlikely to deal with all of them in an unguided manner, e.g. by extracting from the image a full scene graph (Yang *et al.*, 2018), or by classifying and segmenting all the objects in the image (Redmon *et al.*, 2016; He *et al.*, 2017) along with their parts and sub-parts at multiple levels. The desired goal is instead to extract quickly a relevant part of the scene's structure. The relevant part to extract depends on the perceiver's goal, and therefore the visual program applied to the scene should be goal-dependent. An example of applying a BU-TD sequence to a scene in a goal-directed manner is shown in Figure 21 (Guided scene interpretation). The goal in this example was to identify the object that the woman facing the girl is holding.

Figure 21a below, shows the input image. The goal guiding the process is represented by a target structure to find in the image, shown in (b): it is to identify what the woman facing the girl is holding, identifying the object and its properties. Rows (c) to (e) show the sequence of TD instructions used to reach the goal, produced automatically by the algorithm described in Section 4.2 (Composing a visual routine: selecting the next TD instruction). Row (d) shows the sequence of persons extracted from the image, using the instruction <extract-next>. This instruction is used to segment and classify the next person (or scene object) in the image, extracted in order of distance from the camera. The input to this instruction includes a cumulative segmentation map, supplied together with the input image (shown in row c).

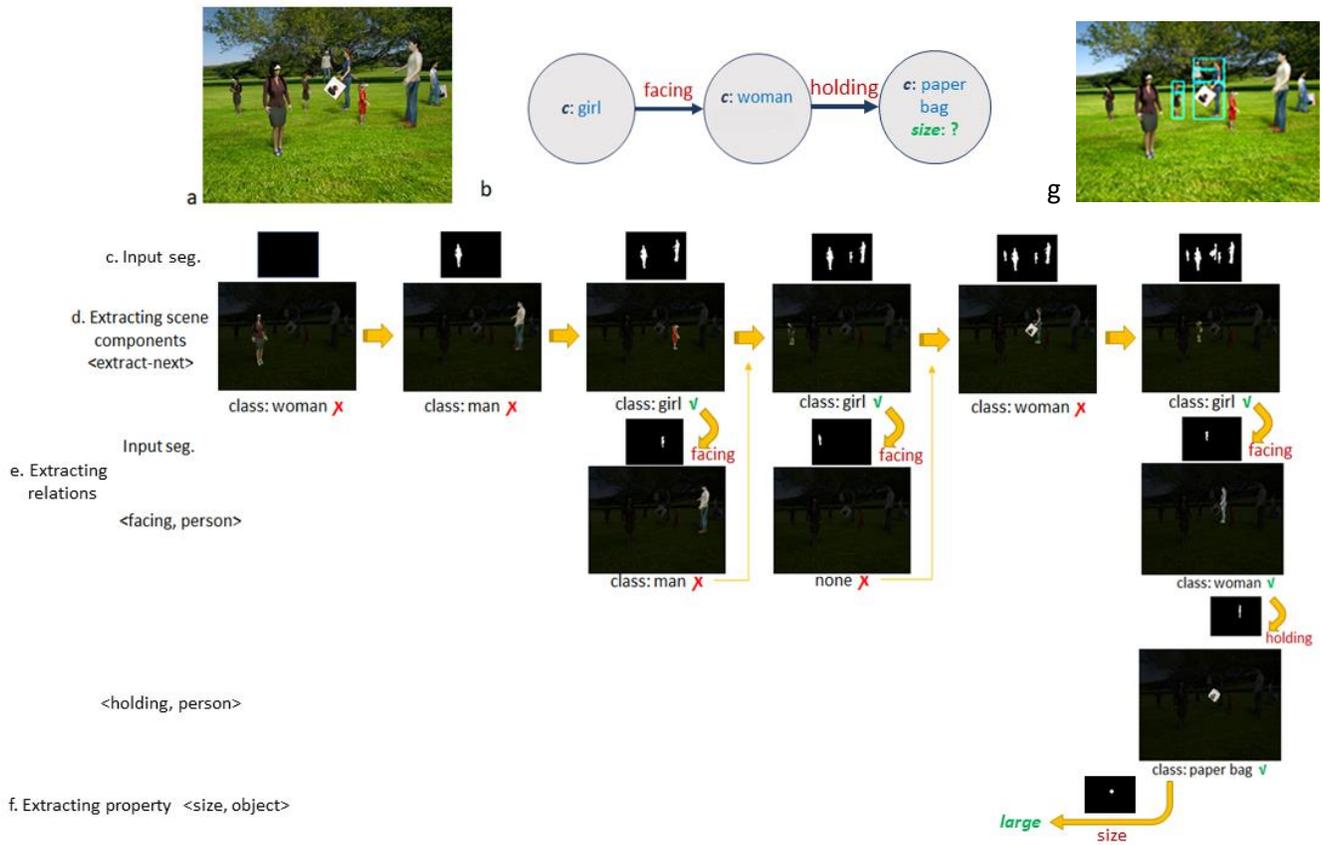

*Figure 21: Guided scene interpretation* The goal is to find a woman who is facing a girl and holding a paper bag, and identify the size of the bag. a. The input image. b. The target structure. Rows (c) to (f) show the sequence of TD instructions used to reach the goal, produced automatically by the algorithm. c,d. Successive extraction of people in the scene, using the TD instruction <extract-next>, which classifies and segments that next scene component in terms of distance from the camera; c shows the cumulative segmentation map of all persons segmented so far. e. Extracting relations, showing the input segmentation map of the reference person (top, smaller image) and the output (bottom, larger image). When the successive extraction by (d) identifies the first girl (3rd image in d), the next TD instruction tests for <facing, person>, producing 'man'. Iterating over d,e continues until a girl-facing-woman is found (last column of d,e). After locating the relevant woman, the next instruction, <holding, person> extracts a 'paper bag' completing the required target structure. In (f) the size of the bag is retrieved by the instruction <size, bag>. g. Illustrates the final result showing the relevant scene components (girl, woman, bag) found during the interpretation process. The final output includes a structured description of the configuration extracted from the image (components, properties and relation, and a grounding of each component in the image, using a segmentation map.

The first person extracted from the image is identified as a woman. The guided search for the structure in (b) is searching for a girl, and therefore the instruction <extract-next> is used iteratively until a girl is found. Following the target graph, the guiding algorithm produces next the instruction <facing, person>, to test whether the girl just discovered is facing another person. The argument to the facing instruction, 'person' is given by the girl's segmentation map, supplied to the network together with the input image. A series of 'next' and 'facing' instructions eventually identifies a woman facing a girl. A <holding, person> instruction verifies that the woman is holding a paper bag and extracts the bag. The goal is reached in (f) by giving a <size, object> instruction and retrieving the size of the bag.

The structure identified in the image is shown in (g), showing the locations of the components (girl, woman, paper bag) in the image. The output of the interpretation process also includes the structural description of the components, relations and properties extracted from the image arranged as an array of components including their related information (properties, relations and segmentation map).

By using a sequence of TD instructions, the process was able to produce a partial analysis of the image and obtain the goal of interest. To achieve the goal, an appropriate sequence of instructions (which is usually not unique) must be applied. A basic question discussed in the next section is, therefore, how an appropriate TD instruction is selected at each stage.

### 4.2 Composing a visual routine: selecting the next TD instruction

In this section, we describe an algorithm for selecting the next TD instruction for performing guided scene interpretation. Scene analysis can also proceed without a specific goal, for example, we can look at a novel image without a specific goal in mind and still proceed to extract useful information. We comment on this unguided extraction briefly below (end of section 4), and focus here on tasks guided by a goal. We assume that the goal used to guide the scene analysis is given in terms of a scene structure to identify in the image, and produce automatically a sequence composed of TD instructions to locate and extract the required structure. In natural situations, the target structure may not be given explicitly, and will have to be constructed. For example, the goal may be given externally by a question stated in natural language, and an initial parsing stage will be required to construct the structure of interest. In the algorithm described below, the process can also conclude that the image does not contain the structure of interest, or identify several (or all) instances of the target structure, in case that the goal requires finding more than a single one. An example of goal-directed scene analysis is the task of visual question answering (VQA), where an image is given together with a specific question, and the goal is to produce an adequate answer based on the image (Kafle and Kanan, 2017). In the VQA task the question, or query, is posed externally, but a query can also be generated internally, by the perceiver, depending on her current goal.

We describe next a general algorithm that takes a target structure and an image as an input, and produces a program composed of TD instructions guided by the target. This process is described in detail in (Vatashsky and Ullman, 2020). In the VQA application, the algorithm first transforms a question, given in natural language, into a graph representation, specifying a scene structure to discover in the image. This stage has to do with processing a question posed in natural language, and will not concern us here. Given a target structure, we focus on the problem of producing an appropriate sequence of TD instructions (which may not be unique), guided by the target. The basic approach for producing a sequence of TD instructions to reach the goal is illustrated schematically using the simple example in Figure 22 (Target-graph). The graph on the left represents the goal of identifying a red car to the right of a yellow bus. To reach the goal and answer the question, a possible sequence of steps is to find a bus, verify its yellow color, find a car to its right, and identify its color. Note that if, for instance, the car is not red or the bus is not yellow, we need to extend the scene analysis process, and search for another car, or another bus. The sequence of TD instructions produced by the algorithm is therefore not a fixed sequence determined by the target structure, but an evolving program that depends on the goal and the image.

The graph on the right in Figure 22 (Target-graph) shows the abstract form of the queried structure. It is abstract in the sense that it does not depend on specific properties and relations: it has an object of class $c_1$ and property $p_1$, in a relation R to a second object of class $c_2$, and a property $p_2$ we need to identify.

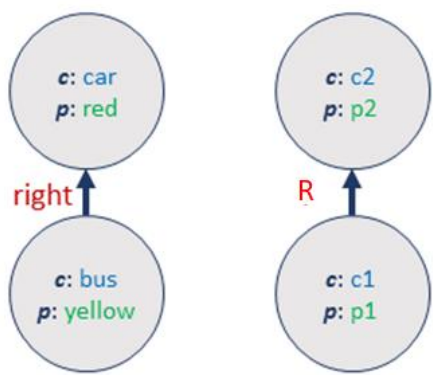

*Figure 22: Target-graph* Left: the target structure to identify in the image. Right: abstract form of the target structure.

Our approach uses the observation that the sequence of steps to follow is determined by the abstract structure of the goal: we can proceed by finding an object of class $c_1$ and property $p_1$, follow the relation R to identify a second object of class $c_2$, and identify the requested property $p_2$. Roughly speaking, the procedure for selecting the appropriate TD instructions proceeds by 'following the graph' of the requested scene structure, selecting instructions to classify objects, extract their properties, follow relations, etc. The procedure we use is also extended to deal with goals that include logical connectives and quantifiers, to find for example all the instances of a given structures, e.g. to address the goal: 'are all the people around the table looking at the speaker?'

The procedure used to produce the TD instructions is a recursive process that handles one node at a time (starting at a root node of a subgraph, if there is one, and an arbitrary node otherwise. The process is carried out by the combination of two BU-TD networks -- an 'expansion network' and an 'elaboration network'. The expansion network extends the representation graph by adding a new scene component (a person or an object) to the extracted scene structure. The elaboration network is used for elaborating the existing graph, by extracting additional properties and certain relations specified by the TD instruction. The two networks are described briefly below, a fuller description is given in Supplementary S7 (Extracting full scene structure). The model comprised of an expansion and elaboration networks was used for extracting scene structures in the Actors data set as described below. The models used for a range of tasks on the Persons and EMNIST data sets (in the sections above and in Section 5 below) used their own networks without using these components.

*The expansion network*: is trained to find a person or an object based on its relation with a reference object, segment it, and recognize its class. For example, the TD instruction to the expansion network can be <facing, person-1>, and the network will then locate the person in the image facing person-1, and produce its classification and segmentation. The argument (person-1) is given to the net by its segmentation map, supplied together with the input image. The network can also extract a new object or person from the image based on its distance from the camera, using the <extract-next> instruction. The argument to the <extract-next> instruction is given by a segmentation map that includes the persons and objects extracted so far from the image.

*The elaboration network*: is trained to recognize a selected property of a specific component according to the TD instruction. The same network is also trained for recognizing relations of the form R(x,y), where the input is composed of two objects and a relation, and the output is the value of the relation (which is often simply True/False value). The input arguments are provided by two segmentation maps that represent the two participating items in the relation. As mentioned above, referring relations between objects (as described in Section 3.3 (Relations)) are extracted by the expansion network, since they are often used to expand the evolving representation.

We outline next the procedure guiding the model to extract the target structure, a more detailed description of the full algorithm is given in Supplementary S9 (Selecting the next TD instruction).
The procedure extracts persons and objects using the expansion network. The persons and scene objects (e.g. a bench or a streetlight) are extracted one by one using the <extract-next> instruction. The algorithm proceeds along the nodes in the target structure, and it guides the model to extract from the image people and objects with their properties and relations that will correspond to the structure described by the target graph. If the class of the extracted person corresponds to the class required by the current node in the graph (e.g. 'girl'), the process next checks the requirements of the node in terms of its properties and relations. The relevant properties are extracted and validated using the appropriate TD property instructions, and the relevant relations are extracted and validated using the expansion network, which retrieves the persons and objects required by the relations in the graph (e.g. 'facing'). Spatial relation instructions (e.g. 'right-of', 'behind') extract people and scene objects, and relations such as 'holding' or 'on' are used to extract tools and objects held by people or placed on other objects. The elaboration network is responsible for relations defined by two arguments (Section 3.3 (Relations)). Property extraction by the network is used either to validate a property in the target graph (e.g., looking for a red object), or retrieving a property (e.g., finding the color of the object on the table). The retrieved information is saved and updated in an array where each entry represents data of one object/person. The retrieved objects that fulfill the requirements are paired with the corresponding objects in the target graph, so that subsequent tests will be applied to the correct objects. Once all the node's tests are completed, the process repeats for the next node, determined by a depth-first (DFS) traversal. The recursive process is terminated either when the full target graph is grounded in the image as required, or when no alternatives are left to check, which may occur already after a partial evaluation of the image, e.g. no objects of a required class were detected. Additional details of the guided process are described in Supplementary S9 (Selecting the next TD instruction).

The guiding algorithm is general in nature, and can extract novel structures of interest in novel scene, by automatically applying appropriated sequences of TD instructions. This capacity to compose and apply visual routines is different from most current approaches (for example, used in VQA), which try to learn the full structure in an end-to-end manner.

### *Unguided scene analysis*

We discussed above the extraction of full-structure, and the guided, goal-directed extraction of a target structure. There is, in addition, a 'default' mode of scene understanding, in which there is no specific goal or a target structure, but we simply look at the scene, and extract some general information of interest. We suggest that in the unguided mode, instead of guidance by a specific goal, the scene analysis is guided at least in part by general priorities, to increase the probability of extracting as early as possible information that is likely to be of interest. In particular, it is generally useful to extract early in the scene analysis process the presence of people in the scene, and discover interactions between people, as well as interactions between people and objects. For instance, to obtain priority for persons, instead of the TD instruction <extract-next> for extracting the next scene component, we can add an instruction <extract-next-person> that extracts the next person in the scene, ignoring scene objects. We currently explore methods for incorporating general priorities in the process of selecting the next TD instruction to guide the model. In addition to such priorities, bottom-up saliency can also affect the order of information extraction from the image, as well as possible other factors, which are left for future studies.

# 5. Capacity and generalization

In the previous section we showed how the BU-TD scheme can be used to extract structures of interest. In the current section we show benefits of this form of task selection in learning a large number of tasks, and generalizing across scene structures.

## 5.1 Multi-tasks and capacity

Successful network models of visual processing focused initially on a single, or a small number of tasks, such as object recognition and segmentation. Recent work started to explore issues that arise when the same model is trained to perform a number of different tasks, such as the recognition and segmentation of multiple objects in the image, multiple properties of the same object, and relations between objects. Important issues that arise when training for multiple tasks are interactions between tasks and capacity limitations, which can make learning more difficult, and reduce the accuracy of the results. The effects can be noticeable already with a small number of tasks, such as in (Sener and Koltun, 2018; Zhao *et al.*, 2018; Strezoski *et al.*, 2019), and they are likely to become increasingly severe, as large scale models of human-like vision will deal with an increasing number of tasks. The issues of multi-tasks and capacity will therefore become critical to their success.

A popular recent approach to the multi-task challenge has been to deal with multiple tasks by using a multi-branch architecture, with a dedicated branch for each task, placed above a common backbone (Figure 23a). Several methods have been proposed for training the multi-branch net, including the use of a weighted loss function (Chen *et al.*, 2018; Sener and Koltun, 2018), or the use of a so-called modulation module (Zhao *et al.*, 2018; Strezoski *et al.*, 2019). One limitation of the branching approach is that it is designed for a fixed number of tasks, rather than dealing with novel tasks, which can be added incrementally at any time. In contrast, the BU-TD model is trained for different tasks using a single common model, trained with different TD instructions. Unlike branching networks and similar models, the BU-TD model does not perform all possible tasks together, but only a selected task, or a small number of tasks, in each cycle. As discussed below, a major advantage of this approach is that it allows to add a new task to the existing model by learning its embedded representation, making it possible to add new tasks without compromising the existing ones.

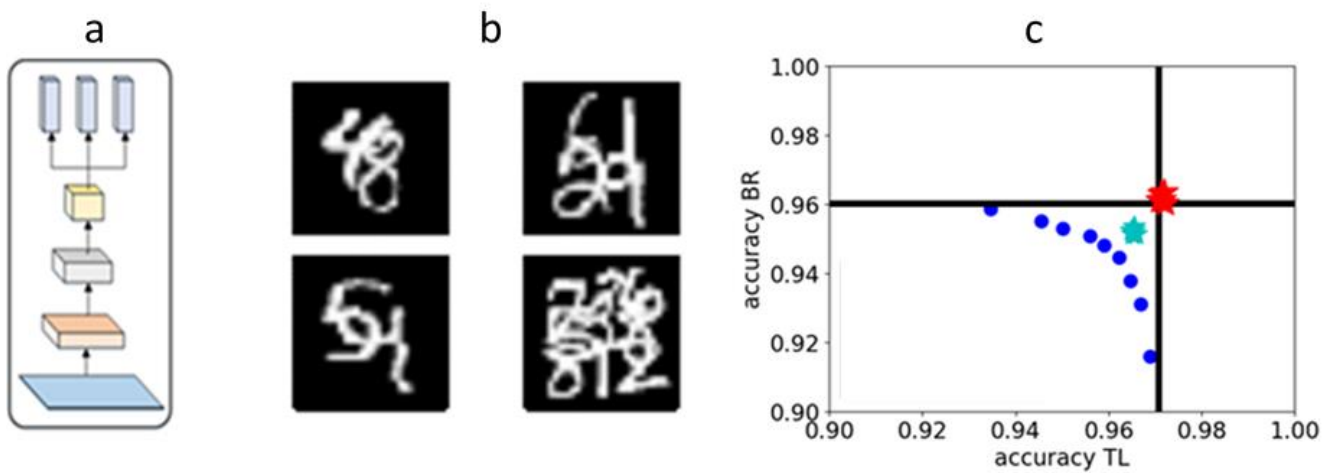

*Figure 23: Multi-MNIST* Multi-tasks applied to Multi-MNIST. a. A schematic branching architecture, of a common backbone followed by multiple task-specific branches. b. Multi-MNIST examples with multiple, partially overlapping digits. c. Performance of a two-digit recognition tasks, x-axis is recognition rate of the digit at the top-left location, y-axis at the bottom-right location. The horizontal

and vertical black bars show performance for each task on its own, blue dots for a branching net using weighted loss, red star for a BU-TD net, cyan star for channel-modulation.

A simple example is illustrated in Figure 23 (Multi-MNIST), using the Multi-MNIST data set (Sabour, S., Frosst, N., and Hinton, 2017; Sener and Koltun, 2018) described briefly in section 3.1 (Objects), in which multiple, partially overlapping MNIST digits are placed in the same image. In the 2-class experiment used in (Sener and Koltun, 2018), the task was to classify the digits in the image top-left (TL) and bottom-right (BR). We added in a similar manner 4 and 9-class experiments, with the digits at each position chosen independently (Figure 23).

We describe briefly the tests and results; full details are described in (Levi and Ullman, 2020). The tests used the MNIST data set with 60,000 examples for each location. We used, as in (Sener and Koltun, 2018), the LeNet network (LeCun *et al.*, 1998) as the common backbone of the tested models. We compared the BU-TD model with several alternatives: a single task baseline, where each task is carried out by its own network, a branching network with a multiple-objective approach (minimizing the weighted sum of the individual losses for different tasks), and a method termed 'channel modulation' (Zhao *et al.*, 2018). Similar to the BU-TD model, channel modulation performs one task at a time, and it uses a channel-wise vector modulation architecture to modulate the weights of different channels according to the task. Performance on two-task experiments is plotted in Figure 23 (Multi-MNIST), where each axis is performance on one task. The blue dots show performance when the two tasks are combined with different loss weights. The plot demonstrates a capacity problem, where higher accuracy in one task is achieved at the expense of reduced accuracy in the other task. In contrast, results of the BU-TD scheme (average of 5 experiments), marked by a red star, shows no decrease, and even slightly better accuracies compared with the single-task case, in which a full network is dedicated to each task. The channel modulation approach (cyan star) shows an intermediate result, with some effect of reduced accuracy for the two tasks. Table 1: Multi Performance summarizes the results for the Multi-MNIST experiment when using 2, 4 and 9 tasks. It shows the average accuracy of all tasks based on 5 experiments for each row. The second column shows the number of parameters as a multiplier of the number of parameter in a standard LeNet architecture. The channel-modulation (extended) model in the table is a channel-modulation model with an increased number of parameters. The use of the BU-TD task selection achieves significantly better results compared with the other approaches, including the single-task baseline that uses 9 networks (probably due to better use of the correlated tasks). Scaling of the number of tasks increases the accuracy gap. It should be noted in the comparisons that in increasing the number of digits, the problem becomes more difficult not only because the number of tasks increases, but also because of the increased overlap between digits, as can be seen in the performance by the single-task networks.

| ALG | 2 digits Av. Acc | 4 digits Av. Acc | 9 digits by loc Av. Acc | 9 digits by ref Av. Acc |
|---|---|---|---|---|
| Single task | 96.46 | 94.15 | 86.62 | 50.33 |
| task-routing | 95.12 | 92.09 | 80.52 | 40.00 |
| channel-modulation | 95.87 | 91.38 | 76.56 | 32.69 |
| channel-modulation (extended) | 96.30 | 92.96 | 79.81 | 38.57 |
| **BU-TD** | **96.67** | **94.64** | **88.07** | **72.25** |

*Table 1: Multi Performance* Accuracy (average of 5 repetitions) on the Multi-MNIST task with 2, 4, and 9 tasks. As the number of tasks increases, the performance gap between the BU-TD and other models increases. As the number of digits increases, each task on its own increases in difficulty as well, due to increased overlap between digits.

*Tasks selectivity*

In the BU-TD scheme, the TD instruction selects a particular task, and the TD stream then guides the next BU pass to perform the selected task. To avoid competition and interference between tasks, a desired goal is to make the processing selective for the selected task compared with the competing, non-selected tasks. To test the selectivity of the representations produced by the BU2 stream, we attempted to predict from the top of the BU2 stream not only the selected task, but also the digits at all locations. For this purpose, following training we attached in the 4-digit test four readout branches (of two layers each) to the top layer of BU2, and trained each of them to predict the digit at one of the locations. The prediction accuracies for the 4 tasks are summarized in Figure 24 (Task selectivity, left panel). The results show over 90% accuracy for the selected task location branch and close to chance-level accuracy for all non-selected branches. Performing a similar readout from the top of the BU1 pass produces accuracies in the range of 30-40% for all locations. We measured task selectivity by a selectivity index, defined as the ratio between (average-accuracy – chance level) of the selected tasks and non-selected tasks. Selectivity index was 26.5 for the 4-task test, and 37.2 for the 9-task test. Additional results and comparisons with alternatives are described in (Levi and Ullman, 2020)

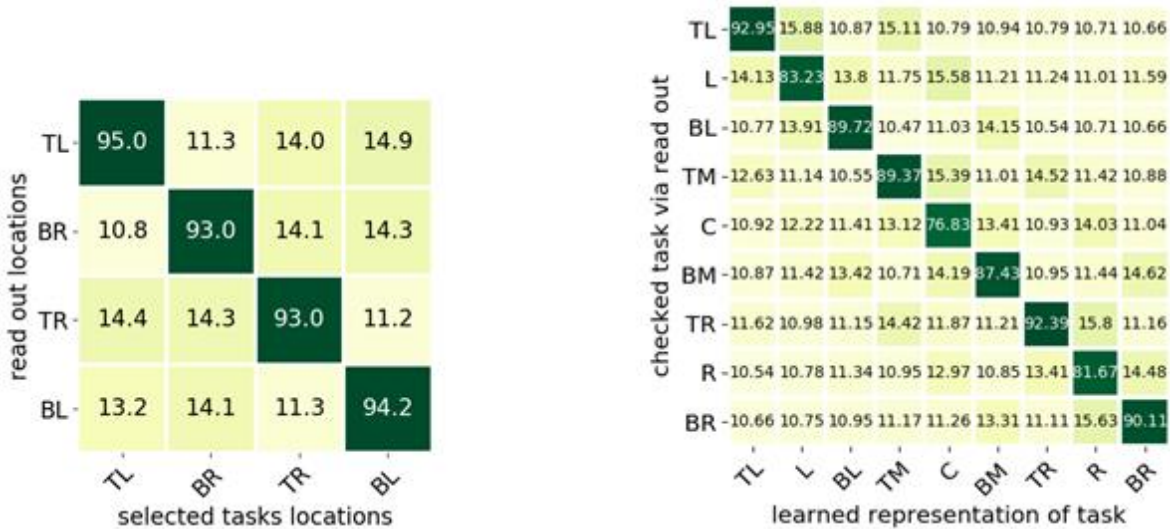

*Figure 24: Task selectivity*   The columns in each plot are the selected task locations, the rows are the readout locations (ranging from Top Left to Bottom Left/Bottom Right). With high selectivity, off-diagonal recognition will be at chance level. Selectivity is shown for the 4-task (left) and 9-task (right) Multi-MNIST task.

We tested task selectivity on a range of additional tasks, with consistent results – the BU-TD model shows high selectivity of the instructed task. For example, in extracting a selected property of a selected person (as in Figure 4 (TD instructions)), the accuracy in reading out the selected property is high (98%), but for reading any of the non-selected properties, the accuracy is close to chance level (this holds for both non-selected properties of the selected person, and properties of a non-selected person).

For EMNIST characters (Section 3.3; Spatial relations), additional tasks we tested were to report the left or right neighbor of one of the selected characters (in a row of 6 characters). The accuracy of reading out the selected neighbor of a selected character was 95%, but the accuracy of reading either the non-instructed neighbor or the neighbor of a non-instructed character were close to chance level.

*Task selection as program modification*

The function of the TD stream in switching between tasks can be considered as a more general form of branching networks. A branching network computes a set of tasks by making use of the fact that the tasks are related in a particular manner: they can be performed by a network using a common part, which is useful for all tasks, followed by a specific branch for each task. This is analogous to a set of programs $T_i$ for the different tasks, which are not independent, but composed of a common part T, followed by an addition part $\Delta_i$, which is specific to each task. Instead of the individual programs ($T_i$), the set of program (T, $\Delta_i$) is an efficient way of dealing with the set of related tasks. The use of TD instructions in the BU-TD model is a more general form of dealing with a set of related tasks. The BU stream can be viewed as a common part T, and the top-down part then modifies T in a task-specific manner. This is analogous to a set of programs $\Delta_i$ (T), where $\Delta_i$ is a program that takes T as an input and produces a modified program for task $T_i$. As in the branching net, this becomes an efficient representation if the tasks are related by a relatively simple modification, which in the general case is not limited to a simple extension performed by branching networks. The results above on task-selectivity show that TD instructions can transform the processing by the BU stream in an efficient manner to focus selectively on the selected task.

In both branching networks and the BU-TD network, the addition of a task adds to the network a new set of parameters (synaptic weights). In the branching net, a new task adds a new branch to the net. In the BU-TD net, the additional task requires learning the weights that create the embedded form of the TD instruction vectors.

The use of branching networks and BU-TD processing can also be combined in a natural manner. The TD counter-stream can be applied to branching networks, and even to tree structures. The branching tree will have a 'counter-tree', and TD instruction will provide inputs to the top of the different branches along the tree. The BU part of each branch in the tree will group together a set of related sub-tasks, controlled by the TD part of the branch. It will be interesting to examine in future studies the ability of such a BU-TD counter-tree structure to provide a general solution to the learning of a large number of tasks, with varying degree of correlations between tasks, together with the ability to add new tasks over time with minimal effect on the previously learned tasks.

## 5.2 Combinatorial Generalization

In extracting structural descriptions from images, we face the problem of dealing with novel configurations of objects, their properties and inter-relations. Even in a limited domain like the Actors data set used in the previous section, a huge number of scenes can be created by selecting different people, their properties, and inter-relations. For natural images the number of configurations is of course much larger, and this large space of possible configurations implies that we need to be able to extract structural representations of novel scenes, based on experience with a small fraction of the set of possible scenes (Tokmakov *et al.*, 2019). The ability to generalize across novel structures, termed combinatorial, or compositional generalization, plays an important role in scene understanding and other cognitive tasks. A recent discussion of combinatorial generalization (Battaglia *et al.*, 2018) concluded that '…combinatorial generalization must be a top priority for AI to achieve human-like abilities, and structured representations and computations are key to realizing this objective.' In the domain of scene interpretation, achieving combinatorial generalization means the ability to deal with scenes with different abstract structures (see Section 4.2 (Composing a visual routine: selecting the next TD instruction)), and for the same structure, the ability to generalize to novel combinations of objects and their properties, as well as to relations applied to new objects.

Combinatorial generalization arises already in simple configurations like the example illustrated in Figure 25a. We used such images in a test where each image contained two persons, each with a number of properties such as hairstyle, glasses-type, shirt, etc. (as in Section 3.1 (Objects)). The task was to recover from the image the identity of the two persons, and the set of properties associated with each of them. We tested aspects of combinatorial generalization in the following manner. A number of person-property combinations were excluded from the training set. For instance, person no. 7 was never presented during training with glasses-type 3. Person-7 was presented with other glasses types, and glasses-3 were shown with different persons, but the particular combination (person-7, glasses-3) was excluded during training. Testing was then divided into two parts: a generalization and a non-generalization test. In the more standard, non-generalization test, the test images showed for at least one of the persons in the image a novel configuration, i.e., compositions of a person, glasses, shirt, hairstyle etc. not seen in training, but without using the excluded combinations. In contrast, in the generalization test, each test image used at least one of the person-property excluded in training. The ability to generalize to such novel person-property pairs is a simple example of combinatorial generalization. Limitation of generalization to such 'excluded pairs' have been noted in modeling network models in both visual and non-visual tasks (Kriete *et al.*, 2013; Lake and Baroni, 2018; Yaari 2018; Bahdanau, Noukhovitch and Courville, 2019; Vankov and Bowers, 2020), but the problem, including tasks similar to the one below, remained largely open.

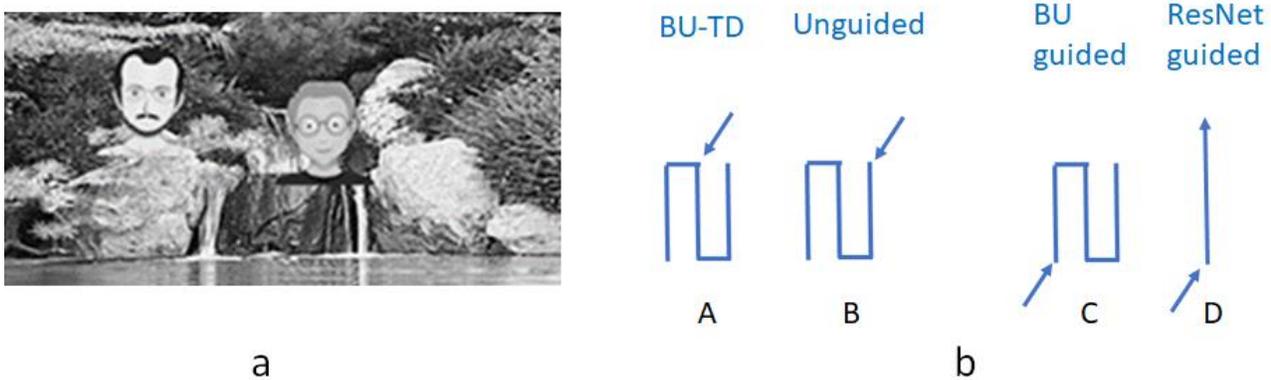

*Figure 25: Combinatorial generalization* a. An example image; the task is to extract the identities of the persons and all of their properties (glasses-type, hair-style etc.). b. Schematic illustration of four network models used for the task. The figure shows the backbone of each model, and the arrow points to the location of providing the instruction to the net. The main comparison is between A, which is the BU-TD guided model, and B, which is a similar network but unguided, producing instead all properties of all persons simultaneously. C,D are versions of guided models, but where the instruction is given together with the input rather than in a TD manner.

We compared generalization in four types of network models, shown schematically in Figure 25 (Combinatorial generalization), described in more detail in Supplementary S10 (Combinatorial generalization). The first model (BU-TD in Figure 25b) is the BU-TD model. The figure shows schematically the BU-TD-BU backbone of the model (as in Figure 3c), and the arrow points to the location of providing the instruction to the net, at the top of the TD stream. Using the instruction, the network applies the selected task, and extracts the required information sequentially. The main alternative is not to use task selection, but to extract all the required information together by a single pass, which is accomplished by the model in (B). To make the comparison between models as close as possible, the model in (B) uses an identical network to the BU-TD model, except for the location of the instruction vector, which is now connected to the last layer of the model (readout selection). In both the BU-TD and readout selection model, the instruction vector may indicate, for instance, <glasses-type, person-3>, in which case the correct output will be the glasses-type of person-3 in the image. The two models share an almost identical structure, and follow the same training procedure. However, unlike the BU-TD model, to provide a correct output to any selected instruction, the readout selection model must extract simultaneously all the properties of the persons in the image in a single pass. The readout selection does not determine which property to extract, but only selects the readout from the last layer. We therefore refer to the readout selection model as 'unguided', as opposed the guided BU-TD model, which is guided to compute the instructed task only.

The selective readout is equivalent in its performance (for a non-generalization test) to a model where all the outputs are read out together, rather than sequentially. We used selective readout in the comparison because it allows us to compare networks with the same structure and the same training procedure, but at the same time one is guided (computing only the instructed output) while the other is not (computing all potential outputs simultaneously). For further comparisons with models producing simultaneous outputs see Supplementary S10.

In addition to the comparison between the guided and unguided models, we also tested two additional variants of the guided model. The model 'BU guided' (Figure 25c) is identical to the BU-TD, except that the instruction is given together with the input image (as an additional input channel), rather than to the TD stream (Supplementary S10). Finally, the model labeled ResNet is a commonly used 'residual network' deep network architecture (He *et al.*, 2017). It has a similar structure, number of units and connections to the BU-TD model, but without the lateral connections between the BU and TD streams. Unlike standard ResNet models, it uses a task instruction, similar to the BU-TD network.

We consider next combinatorial generalization results of the different models. The central comparison is between the first two models, comparing the BU-TD model with an unguided model that produces all the properties together. All models were first trained on a data set of training images, and tested on a test set images, without the excluded combinations, to perform the non-generalization test described above. The models were next tested on the excluded pairs, and the results are reported in the generalization test. For each task, we used two training set sizes: a small size and a large size. The small size was the smallest set of training examples for which the BU-TD achieved an accuracy criterion of at least 85% on the excluded set (labeled below as 'sufficient set'); the large was several times larger (labeled extended set, see figures). Results for combinatorial generalization and non-generalization tests are shown in Figure 26 (Person-property generalization). The top row of Figure 26 shows the accuracy of extracted properties on the sufficient set, with a training set of 1600 images. The training of each model continued over multiple epochs until a pre-set convergence criterion (an improvement of less than 2% over 50 epochs) was reached (see training details in Supplementary S10). The bottom row shows the results of the same models on the extended set. Our main comparison is between the guided BU-TD model (labeled A in the figure, green) and the unguided model (B, blue). As can be seen, the BU-TD model reaches high accuracy already with the sufficient set, and the accuracy is high for both the non-generalization and the generalization tests. In contrast, the unguided model (B) eventually reaches high accuracy on the non-generalization test (extended set, bottom row), but the accuracy in combinatorial generalization remains low, close to chance level. (23%). The alternative guided models (C,D) require longer training, but they all eventually reach high generalization accuracy.

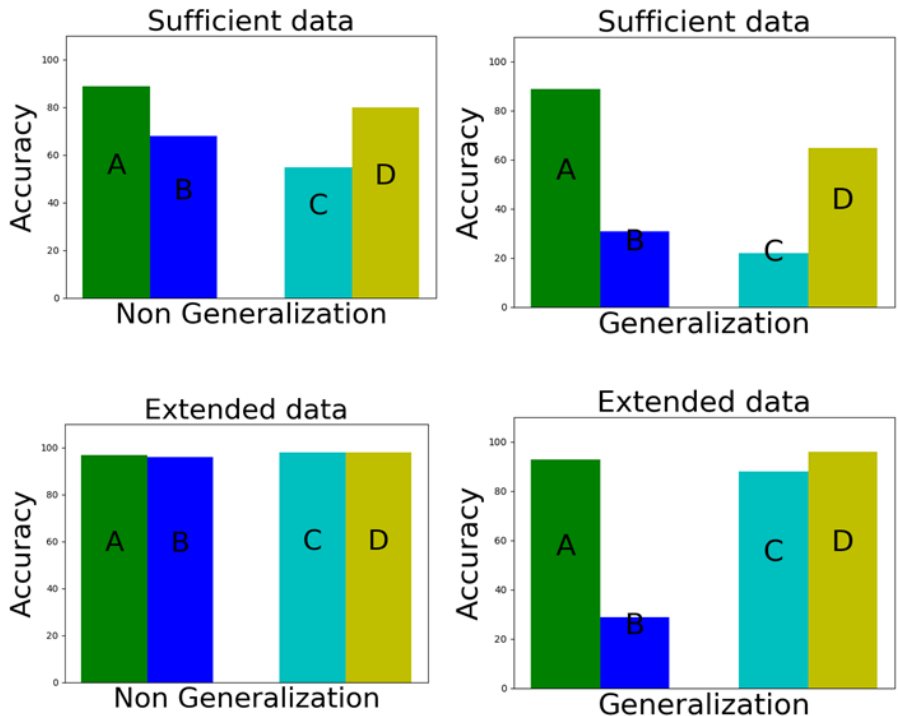

*Figure 26: Person-property generalization*  Combinatorial generalization, person properties. Top row: training with 1600 examples (sufficient data) until convergence.  Bottom row: training with 4,800 examples (extended set) until convergence. Results (average accuracy) are for the four models in Figure 25. The main comparison is between the guided BU-TD model (labeled A), and the similar, but unguided version (B).  The models in C,D are alternative guided models. For combinatorial generalization test, the unguided model remains low, close to change level (23%). With extended data (bottom), all guided models exhibit combinatorial generalization, while the unguided model remains low.

### *Combinatorial Generalization:  right- and left-neighbor relation*

We performed similar tests of combinatorial generalization with a different task, shown in Figure 27 (Relation generalization), generalizing the spatial relations of left-of and right-of. Images for this test contained a set of 24 EMNIST characters, selected from a total of 29. The task was to produce the right-neighbor and left-neighbor of each of the characters. The BU-TD network was trained for this task with two losses. One (binary cross entropy loss) at the end of the first BU pass that measures the accuracy of identifying all the 24 characters present in the image. Second, the model receives an instruction of the form <direction-of, x> where x was one of the characters in the image, and it was required to produce the right-neighbor or left-neighbor of x at the end of the second BU pass (cross-entropy loss). The networks used for comparison were the same as in the previous example (Figure 25 (Combinatorial generalization)).

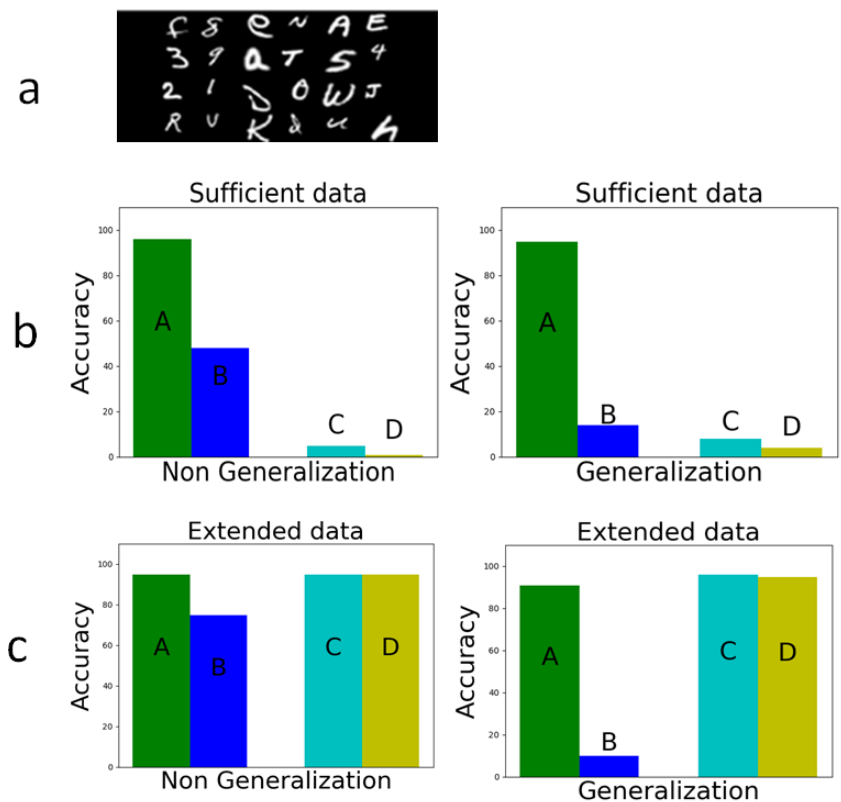

*Figure 27: Relation generalization* Combinatorial generalization for EMNIST right-of, left-of task. a. Example of an input image, with 24 characters. b. Results (average accuracy) for training with 100,000 character-pair examples (sufficient data). c. Results for the extended training set of 400,000 character-pair examples. The models tested are the same as in Figure 26 (Person-property generalization). The main comparison is between the guided BU-TD model (labeled A), and the similar, but unguided version (B). The models in C,D are alternative guided models. For the smaller number of examples, only the BU-TD model achieve high accuracy on both non-generalization and combinatorial generalization tests. With sufficient data, all the guided models obtain combinatorial generalization, the unguided model fails to generalize.

Training was done as before on two data sets. The smaller (sufficient) set was trained on 10,000 images, selecting 5 character at random from each image for training and both directions, for a total of 100,000 training instances (character-pairs). The larger (extended) set used all 20 pairs from each image and both directions, for a total of 400,000 character-pairs. One set of 24 characters was used as the excluded set: all pairs of successive characters in this set were not shown in any of the training images.

Similar to the previous example, by the time the BU-TD model learns the task (the sufficient data set), the alternative models produce lower accuracies in both the generalization and non-generalization tests (Figure 27b). When the number of training examples increased (to 400,000), the instructed models catch up and achieve good generalization, but the BU model with no instruction again shows little or no generalization (Figure 27c). We repeated this experiments with a larger fraction of excluded pairs (up to 59% of all pairs), and combinatorial generalization obtained by the BU-TD model remained the same as above. Additional details of this test are given in Supplementary S10 (Combinatorial generalization).

In the results above, the unguided model (as well as other models without task-selection, see Supplementary S10) showed no or marginal generalization to combinations not seen in training. We further tested whether increased generalization may emerge in the unguided model by sufficiently simplifying the task.

As shown in Figure 28 (EMNIST-6), results showed that generalization became possible when the EMNIST task was reduced to 6 characters, and using a large number of training examples. Figure 28 shows the accuracy results after training with the sufficient set of 2,000 test images, (with 10,000 pairs). As before, the BU-TD reaches high accuracy, while the unguided model remained low. With a large set of training images (10,000 with 50,000 pairs), combinatorial generalization accuracy of the unguided model improves and approaches its accuracy for the non-generalization case, but remains consistently lower that the BU-TD model (77% vs 94%). The learning was slow and required about 10 times more presentations of the examples (iterating the examples for more epochs) compared with the guided model. Finally, we repeated the test, but making combinatorial generalization more challenging by increasing the percentage of excluded digit-relation pairs during training. As can be seen in Figure 28d, accuracy of the unguided model decreases sharply, dropping to 3% at 63% of excluded pairs.

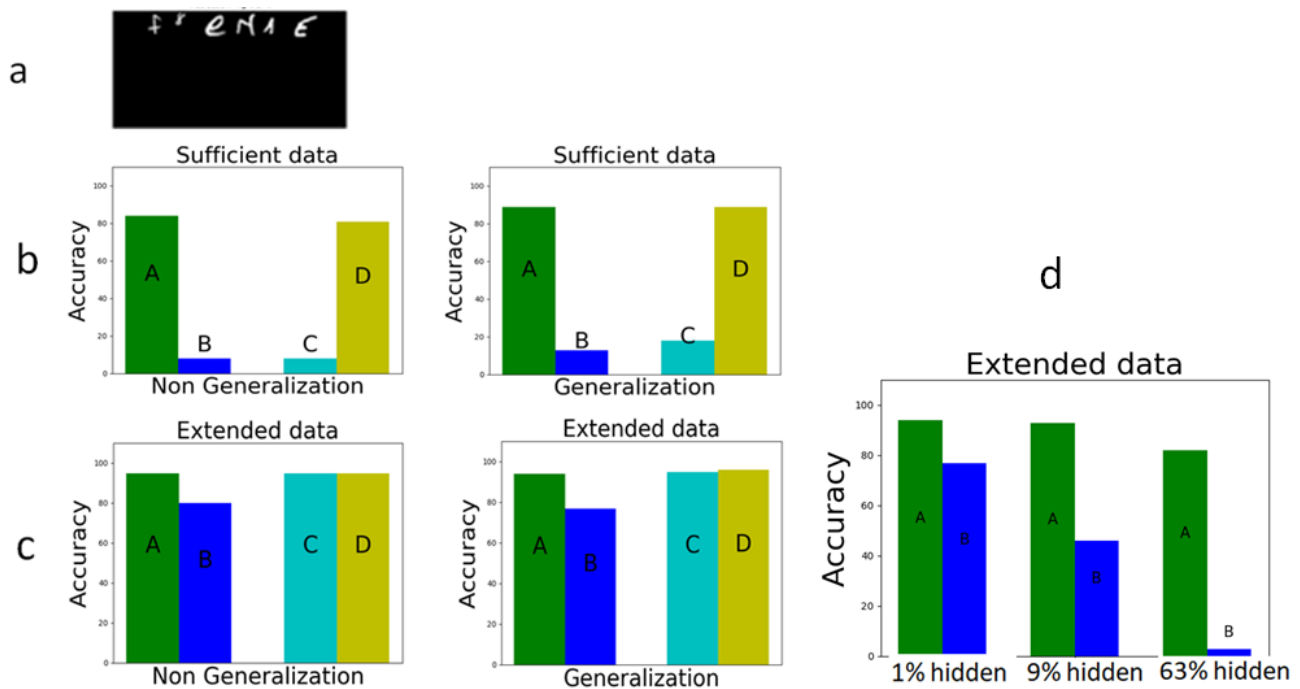

*Figure 28: EMNIST-6 generalization* Combinatorial generalization for a 6-character EMNIST right-of and left-of tasks. a. Example of an input image. b. Results after training with 10,000 character-pairs(sufficient set); the BU-TD model reaches high accuracy, the unguided model remains low. With an extended set of training images (50,000 character-pairs), the unguided model achieves significant combinatorial generalization, but saturates at a lower level than the unguided models (77% vs 94%). d. Increasing the percentage of character-pairs excluded during training. Accuracy of the unguided model drops to 3% at 63% of excluded pairs.

These results show that achieving some measure of combinatorial generalization in the unguided model is possible when the task is sufficiently simplified and the number of training examples is increased. In contrast, it is worth noting that in some common models of learning multiple tasks simultaneously (such as multi-branch networks discussed above), combinatorial generalization cannot take place due to limitations placed by the specific architectures used by the models (Supplementary S10).

In summary, the generalization tests have two main conclusions regrading combinatorial generalization. First, the instructed models produced a much better generalization compared with the non-instructed model, which extracted all the properties or relations simultaneously. Second, among the alternative instructed models, the use of instruction by the BU-TD showed an advantage compared with the alternatives, in terms of generalization performance, number of examples and overall number of presentations required for learning.

*Combining generalization with compound instructions*

Finally, we also tested the combination of generalization with the use of compound instructions, discussed in Section 3.1 (Objects, subsection Properties, compound instructions). In a compound instruction, the BU-TD model is trained on multiple tasks, where each task is trained on its own; for example, extracting a single property of a person or object. Following training, instructions for multiple properties are combined into a single instruction, and they are extracted simultaneously rather than sequentially. The question we consider here concerns the effect of compound instructions on generalization, whether the tasks learned separately but performed together, will show a similar combinatorial generalization, compared with tasks that are learned and performed separately, as in the previous sections. As above, we used for this test images showing two persons, with the task of extracting their individual properties. The BU-TD model was trained using TD instructions to extract at each cycle a single property of one of the persons in the image. During test, the TD instructions to extract two properties of the same person were combined into a single instruction, thereby extracting the two properties within a single cycle rather than sequentially. To test combinatorial generalization, some person-property combinations (e.g. person-2, glasses-type 3), were excluded again during training. We then combined two tasks into a single instruction, where one of the two made use of an excluded combinations. Similar to the tests above, the same accuracies were obtained by the BU-TD model for the generalization and non-generalization tests. The generalization results for compound instructions raise the possibility of combining advantages arising from sequential learning, with advantages of parallel execution. Different properties can be learned independently, and at different times, but they can be later recovered simultaneously, without requiring additional training, and without detracting combinatorial generalization.

### 5.3 From symbolic to embedded instruction representation

A general aspect of TD instructions is that they can be represented initially, at the high-level parts of the system, in a symbolic form, and through learning they transform to an embedded form that is useful for guiding the subsequent processing of the visual stream.

In a symbolic system, there is a mapping between elements of one domain, used for representing a second, so-called 'semantic' domain; for instance, words representing objects and other entities in the world. In the way used here, the 'symbolic' (or 'purely' symbolic) representation carries no information about the object being represented. In an embedded representation, the symbols are embedded in a vector space in such a way that the vector representation of an object carries information about the represented object.

In this section we describe aspects of the transition from a purely symbolic to an embedded form in the TD instructions guiding the BU-TD model (Section 2.2 (Learning in the counter-streams model) and Figure 6 (Providing instruction)). In most of the experiments we conducted, the TD instruction provided to the network was already divided into two components – a task to perform, and an argument to apply the task to, such as extracting a property of a selected person or object. In the experiment below, the full instruction was provided instead by a single one-hot vector (combining task and argument), and we found that the compositional form was acquired on its own in the learned embedded representation. Furthermore, *the ability of the model to reach combinatorial generalization depended on the emergence of the compositional embedded representation.*

An example is the extraction of selected properties from a person as in Figure 4 (TD instructions), Section 2.1 (Network structure). We used a dataset of images with two persons per image, selected from a set of six persons, with five different properties. There was therefore a total of 6*5 = 30 possible tasks, where a task is, e.g. identify the shirt-type of person-4. Each of the tasks was represented by a one-hot vector, namely, a 30-long vector with a single '1' at the appropriate position. The instruction vector then creates an embedded form as in Section 2.2 (Learning in the counter-streams model) and Figure 6 (Providing instruction), except that the original instruction was not divided into separate task and argument parts. That is, instead of using $I_{task}$, $I_{arg}$, for the task and argument, we used a single one-hot instruction vector $I$. As before, the vector $I$ is then transformed into a learned embedded form $E$ which provides an input to the TD stream.

The model was then trained on the different tasks. As above, we tested combinatorial generalization, by excluding from training a certain fraction of the tasks, with two questions in mind. First, comparing the model performance using the single symbolic instruction form, against the compositional form, in which the TD instruction was divided into an object and a property. Second, when the tasks are given initially in the non-compositional form, will the compositional form emerge on its own in learning the embedded representation? To test the second question, we performed a readout from the embedded representation after learning. The readout (two linear layers with a ReLU) was trained to recover the selected person and property in the instruction, using some of the instructions during training and testing on the rest. For example, if one of the one-hot instruction vectors specifies the combination: (glasses-type, person-2), will it be possible to recover the person and the property 'glasses-type' from the embedded instruction, without seeing the instruction during readout training? Such a readout will not be possible from the symbolic representation, but may arise in the embedded form during learning, indicating a learning of the compositional representations.

In order to test the model performance we started with a simple version of the learning task, in which a small proportion (7.5%) of the possible pairs (such as glasses-type-4, person-2) where excluded from training. The non-compositional model learned the tasks well, achieving 91% combinatorial generalization accuracy, compared with 93% using the standard instructions used explicitly the compositional form ($I_{task}$ $I_{arg}$). In learning to read out the person and property from the embedded form, test accuracy was 100%. In this case, the model learned the task well, and produced on its own a transition to a compositional representation.

Learning was then made increasingly difficult by excluding a low (7.5%), medium (40%) or high (75%) fraction of person-property pairs. In the 'high' condition, (75%), the exclusion was not just leaving out more pairs, but excluding from training a complete property, e.g. the instructions for (person-2, glasses-type-k) were excluded from training for all values of k, so that person-2 was never trained with the glass-type property, and the entire task (glasses-type, person-2) was not used in training. A total of 9 out of the 30 tasks were excluded during training.

| Exclusion | Readout accuracy | Prediction accuracy | Compositional instruction | |
|---|---|---|---|---|
| Low (7.5%) | 100 | 91 | 93 | 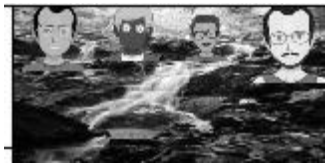 |
| Medium (40%) | 68 | 62 | 91 | |
| High (75%) | 9 | 32 | 80 | |

*Table 2: Symbolic-embedded* From symbolic to compositional representation of the TD instruction. The task was to extract a selected property from a selected person (example image on the right). 'Readout accuracy' column is the accuracy (in percent) of reading out the instructed person and property from the embedded instruction representation; high values indicate successful readout of the compositional structure. 'Prediction accuracy' is the accuracy of combinatorial generalization. 'Compositional instruction' shows prediction accuracy when the TD instruction is given in the standard compositional form (using two instruction vectors). The results show that combinatorial generalization (prediction accuracy) arises together with the emergence of compositional representation (readout accuracy).

In order to answer the second question (emergence of compositional representation) we performed a readout from the embedded representation after learning for each of the three conditions. In all conditions, when training the readout task the exclusion was a complete property, e.g. when reading out the instruction: <glasses-of, person-2> we exclude during readout training all examples of person-2 with any type of glasses. The combination of person-2 with the property glasses-of was therefore novel at readout test time.

The results are summarized in Table 2: Symbolic-embedded. The table has 3 rows for the different conditions – excluding in training a low, medium or high proportion of the person-property combinations. The first two columns show two measures for each condition: the readout accuracy of the instructed person and property from the embedded representation, and the accuracy of the model in testing combinatorial generalization. For comparison, the third column, 'compositional instruction', shows accuracy when using the compositional form of the TD instruction.

As shown by the table, when the instruction is given in a non-compositional form (by a single one-hot vector), the two measures go down hand-in-hand: the prediction accuracy goes down, and at the same time, readout of the person and property from the embedded representation also deteriorates. In addition, the symbolic instruction also required a significantly longer training (e.g. 348 vs. 148 epochs for 40% exclusion).

In the original BU-TD model, the TD instruction was given already in a compositional form: the instructed person and property to extract are given separately ($I_{arg}$, $I_{task}$ in Figure 6 (Providing instruction)), each by its own one-hot vector. As the difficulty increases, the accuracy of the model decreases as well, but remains higher: 93%, 91%, 80% for the low, medium and high exclusion conditions, respectively.

The results indicate that the compositional form of the TD instruction was crucial for combinatorial generalization. When the TD instruction was provided in a symbolic one-hot form, generalization was possible only as long as the components, the person and the property, could be read out from the embedded task representation. Such advantage of compositional representation may be expected, since combinatorial generalization depends on the ability to deal with new combinations of existing components. For the easy task (training with almost all possible combinations), the transition from the one-hot to a compositional representation appeared on its own during learning. However, when the task became more difficult, the transition failed, and the components could not be read out of the embedded representation. Implications of this finding and their relations to innate structures using compositionality are discussed further below in the final discussion.

# 6. Relations to human vision

This section discusses briefly some relations between the model and human vision, with possible directions for further future studies.

## 6.1 Counter-streams in primate vision

A prominent difference between feed-forward deep networks and the primate visual cortex is that in the visual cortex there are massive top-down connections going from high to low levels of the hierarchy of cortical regions. The connectivity between visual areas is typically reciprocal, that is, if visual area *A* sends its outputs to a higher-level area *B*, then there are also reciprocal connections, going down from area *B* to area *A* (Van Essen and Maunsell, 1983; Markov *et al.*, 2013). Based on anatomical and physiological data, a model of the BU and TD flow in the visual cortex (Ullman, 1995) proposed the schematic connectivity diagram shown in Figure 29 (Counter streams cortical structure), with extensions based on (Markov *et al.*, 2014). The figure shows the connectivity between two interconnected cortical area (I) and a higher-level area (II). The BU stream goes through layers 4 and 3B of area I to layer 4 of the higher area. The TD path goes through layer 2/3A and layer 6 of area II to 2/3A and 6 of the lower area. The separation in the superficial layers between an ascending and descending populations was predicted in (Ullman, 1995), and demonstrated in (Markov *et al.*, 2014).

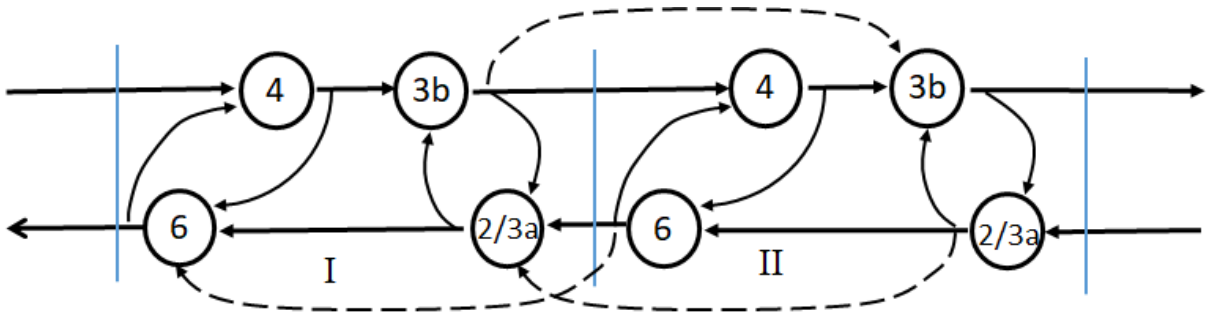

*Figure 29: Counter streams cortical structure* (adapted from (Ullman, 1995), additions from (Markov *et al.*, 2014)). How the basic counter-streams structure may be embodied in cortical connectivity of the ventral stream. The main connections between successive areas labeled I, II. The ascending BU path goes through layer 4 to layer 3B of area I and then to the next area. The descending TD path goes from layer 2/3A and layer 6 of area II to 2/3A and 6 of the lower area. The two streams are interconnected in both directions, between layers 4 and 6, and within the superficial layers 2, 3. Dashed arrows are pathways that skip a step in the stream.

As noted in section 5.1 (Multi-tasks and capacity), the counter-streams structure can also be extended to a counter-tree form in which TD instructions provide inputs to the top of the different branches along the tree. A branch in the tree will group together a set of related tasks, controlled by the TD part of the branch. In the primate visual system, there is evidence suggesting some form of branching of the visual streams into major categories such as faces, objects, and places and possibly sub-divisions such as natural and man-made objects (Chao and Martin, 2000), or indoor and outdoor scenes (Henderson *et al.*, 2007). It will be of interest to explore further BU-TD processing in such hierarchical tree structures, both computationally and in modeling the primate visual system.

### 6.2 Symbolic and compositional /embedded representations

In section 5.3 (From symbolic to embedded instruction representation) we compared the use of 'symbolic' and 'compositional' TD instructions. In the compositional form, the task to perform, and what to apply it to, are given separately, and in the 'symbolic' form they are combined in a single instruction, e.g. in a single one-hot vector (Zhao *et al.*, 2018). The results showed a clear advantage of the compositional form. They also showed that the compositional form can be learned during training, but that starting training from a compositional form has a substantial advantage in terms of combinatorial generalization.

Given the importance of combinatorial generalization, it appears highly advantageous to use TD instructions in their compositional form from the start of learning different visual tasks, rather than learning such representations for each task separately. The use of compositional representations in the BU-TD model is only an example of the broader use of compositional representation in perception and cognition, but it can serve to study further issues related to innate vs. learned representation (Ullman, 2019). There are two general possibilities (and intermediate options in between). One is that TD instructions are given in a non-compositional form, and the compositional structure is learned for each task on its own. A second possibility is that the advantage of a compositional representation in guiding the visual stream was discovered throughout evolution, and has some innate aspects which makes learning more efficient. From the viewpoint of human perception and cognition, it will be of interest to study combinatorial generalization in visual perception during early development to try to identify its innate and learned aspects. From the viewpoint of computer vision, it will be of interest to explore the use of incorporating compositional representations into the structure of network models, and study their effect on learning visual tasks in general, and on combinatorial generalization in particular.

### 6.3 Functions and benefits of TD guidance

The main function of the TD stream in the model is selection as well as guidance of what to perform next. Using 'cognitive' instructions, it becomes possible to select a task to perform and an argument to apply the task to, and the TD stream can then guide the task in the sense of modifying the BU stream to perform the selected task. To obtain this function the model is trained with a combination of the input image together with and a corresponding instruction.

Studies of the BU-TD structure have shown a number of related benefits for the purpose of extracting relevant scene structures. Two useful outcomes discussed in Section 5 (Capacity and generalization) are combinatorial generalization and dealing with capacity limitations. In terms of capacity, the main contribution of the BU-TD structure is that an embedded task representation is learned for each individual task. As a result, each task has its own set of embedding parameters, adding to the parameters of the main BU-TD backbone, thereby contributing to the overall capacity.

In theories of human perception, capacity limitations are linked to the use of early selection by attentional processes. The general notion is that visual scenes typically contain more items than can be processed at any one time due to the limited processing capacity of the visual system. Selection processes are therefore used to deal with this capacity problem by selecting relevant information (Van Essen and Maunsell, 1983; Maunsell, 2015) . For example, the role of visual attention is defined in the Encyclopedia of Neuroscience (McMains and Kastner, 2008) as: '*Visual attention* refers to the cognitive operations that allow us to efficiently deal with this capacity problem by selecting relevant information and by filtering out irrelevant information'. The studies of Section 5 (Capacity and generalization) show for the first time another major function for the use of early selection, in terms of affording combinatorial generalization. Finally, as noted in the introduction, the use of 'cognitive' instructions also helps guiding the scene analysis process towards the extraction of relevant information. To obtain this benefit, we suggested that the scene interpretation will require a 'cognitive augmentation' stage on the cognitive side of the scheme, which was not discussed in the current work, but will be of interest to explore in the future. Finally, as discussed further in Section 4 (Extracting scene structures using BU-TD sequences), by the use of multiple BU-TD cycles guided by a sequence of TD instructions, it becomes possible to perform complex visual tasks, in particular the extraction of structural descriptions of interest. It also becomes possible to produce new compositions of existing TD instructions, creating new 'visual programs', and in this manner make 'infinite use of finite means' (von Humboldt, 1836; Chomsky, 1965) to perform novel visual tasks.

# Supplementary

**Contents**

## S1 The psychophysical timeline study

In an ongoing study (S. Ullman, D. Harari, H. Benoni) we tracked the extraction of structured information (scene components, properties, relations) from images over time (50-2000 msecs). The study was conducted online with subjects recruited via Amazon's Mechanical Turk platform. Each image in the study was presented to a subject in one of 7 presentation times (50, 75, 100, 125, 200, 500, 2000 msecs.) followed by a mask. Each subject saw a given image presented at only one of these time durations, and each image, at any of the 7 time durations, was seen by 25 subjects. The subjects were asked to produce following each presentation a detailed description of the scene in the image, listing all objects, people, their properties and interactions. The figures below show examples of interpretation timelines. The first shows the timeline of the 'stealing' scene discussed in Section 1 (General background and goals: combining vision and cognition), in discussing the integration of vision and cognition. The second shows two images discussed in Section 3.3 (Relations), in describing referring relations.

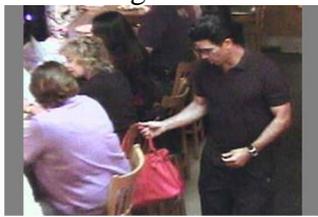

- purse belongs to the woman
- bag is on or hanging of chair
- man stealing a purse
- man stealing from woman

- man standing
- chair mentioned
- restaurant / crowded place

- woman is sitting
- man wearing black
- shirt mentioned

- bag / purse mentioned
- man reaching towards the woman

- man and a woman
- man is reaching

- 1 or 2 persons

Time, msecs

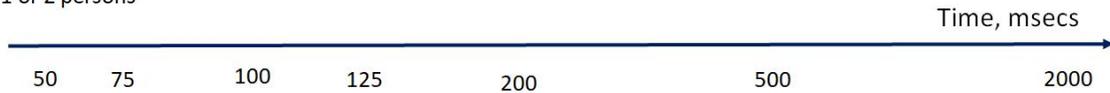

| 50 | 75 | 100 | 125 | 200 | 500 | 2000 |

*Figure 30: Stealing timeline* The extraction of scene components, their properties and relations over time. The figure shows for each time bin the information added to the scene description by at least 5 out of 25 observers.

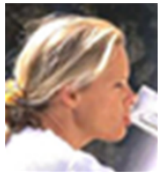

drinking

From a glass
Has pony tail

Blond woman facing right

Time, millisecs

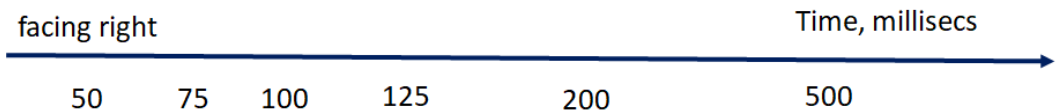

| 50 | 75 | 100 | 125 | 200 | 500 |

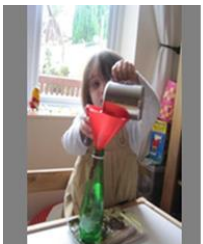

Into a bottle

Through a funnel

Is pouring

Holding object

A girl/child

Time: milliseconds

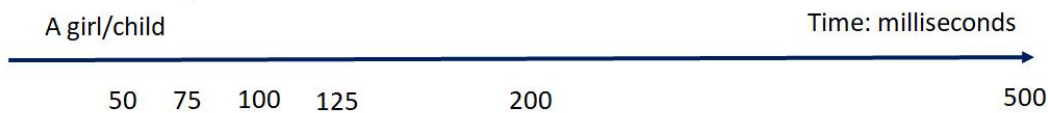

| 50 | 75 | 100 | 125 | 200 | 500 |

*Figure 31: Relations timeline* The figures shows for each time bin the information added to the scene description (by at least 5 out of 25 observers). In these images of a person engages in an action, the person and action are recognized before the object used in the action.

## S2 The BU-TD structure

The BU-TD counter-streams model is a general scheme, in the sense that BU-TD models can be genrated from different BU models, by adding the TD counter-stream and the lateral connections. The main structure we used was based on a bottom-up Resenet 18 network, but we also used both shallower and deeper network models for different tasks, such that a BU-TD network for tasks related to the Persons data set, the EMNIST data set, and the Actors data set. A single BU-TD network can be trained on multiple different tasks, but it is also possible to use more than a single network.

The TD instruction provided to the model is a composition of <task, argument> pair. For example <right-of, 5> or <left-of, w> for the EMNIST and <glasses, person-1> or <mustache, person-5> for the Persons task. The task and argument are encoded separately as a one hot vector each, and then transformed to an embedded form by learned weights. This embedded vector provides an input to the top layer of the TD stream.

*Cross-streams lateral connections:* The BU-TD scheme includes two sets of cross-stream lateral connections, one in the BU→TD direction, and the other in the TD→BU direction. We used in our experiments two alternative schemes for the lateral connections, based on computational comparisons among a larger number of alternative versions. One of the alternatives uses additive lateral connections, similar to the connections used in standard deep network models. In the second alternative used for the TD→BU direction, we applied a multiplicative interaction described below.

Consider a unit $\bar{x}_{i,k}$, which is in channel i and location k of a layer on the TD stream. (An upper-bar over a variable, such as $\bar{x}$, is used to denote variables on the TD stream.) The total input I to this unit is composed of $\bar{I}_S + I_L$, where $\bar{I}_S$ is the standard input along the TD stream, and $I_L$ is the lateral contribution from the BU stream. One form for $I_L$ we used was convolutional, using: $I_{L_k} = \Sigma_c \, w_{i,c} x_{c,k}$. That is, the lateral input to $\bar{x}_{i,k}$ on the TD stream comes from all the units $x_{c,k}$ at the corresponding location in the corresponding layer on the BU stream, across all channels. The weights are independent of the location k (also called '1*1 convolution' on the BU side). The weights do depend, however, on the target channel i. In a simplified case, we used a similar version where the inputs to a target channel i on the TD stream came only from the single channel i on the BU stream, rather than all channels. In the multiplicative scheme (also called a 'gating' scheme), the lateral contribution $I_L$ uses the convolutional form $\Sigma_c \, w_{i,c} x_{c,k}$, but it is used in a multiplicative form: $\bar{x}_{i,k} = \bar{I}_S * I_L = \bar{I}_S * \Sigma_c \, w_{i,c} x_{c,k}$. This form can also be combined with an additive term, $\bar{x}_{i,k} = \bar{I}_S * I_L + \alpha I_L$ (with $\alpha$ a learned parameter).

*Training multi-cycles:* The BU-TD model is a recurrent network, which can be used for multiple cycles. To train the network, the unfolded network is used to train a single cycle, of BU, TD, BU (Section 2.1 (Network structure)). To train a process composed of multiple cycles, we used two options. The first is using an unfolded network for several time steps. The second is to train a large number of cycles using two networks, one for the BU and the other for the TD part. This is accomplished by storing and passing activation values between the two networks. Activation values of the BU streams are recorded, and used as inputs to the TD network on its next cycle, and similarly activation levels of the TD stream are recorded and used as input to the BU network in training its next cycle.

*Using intermediate losses*: The BU-TD network can naturally use intermediate outputs and losses at the end of each stream. Although a single loss at the top of BU2 is often sufficient for learning, the use of intermediate losses is useful in terms of accelerating the learning process, and providing useful additional outputs such as segmentation maps of detected objects at the bottom of the TD stream. For example, we experimented with a different number of losses for the EMNIST dataset. The task was computing spatial relations in images containing 24 characters (as in Section 3.3 (Relations), Figure 14 (Spatial-2D)). The network was trained from initial random weights, with two losses. At the top of BU1, the net was required to identify all the characters present in the image, using binary cross entropy. The second loss was at the final output, producing a single character, using a cross entropy loss. The training set included 10,000 images, and in each training epoch, each image was shown 10 times with 10 different TD instructions, for a total of 100,000 single-character outputs. Under these conditions, performance reached 94% accuracy at 33 training epochs (asymptoting at 96%, 39 epochs). Varying the number of losses affected the accuracy and learning speed: using a single loss, end-to-end training failed for this task, asymptoting at only 4% accuracy. Using three losses, an additional segmentation loss of the instructed character at the end of the TD stream, achieved 94% accuracy in 13 rather than 33 epochs. In other tasks we tested a single final loss was often sufficient, but at the expense of longer training.

*Using pre-trained models:* In training for multiple tasks, the tasks may be acquired sequentially in some natural order. For example, recognizing individual objects may be learned before learning to recognize spatial relations between objects. Training on a task T may become easier, if T is learned in a network already trained for related tasks. As an example, we compared training a model on the task of spatial relations (right-of, left-of) between EMNIST characters either on its own, from random weights, or starting from a network already trained on individual character recognition. For training on the character recognition task, the BU-TD network was given images of 24 characters, with an instruction of an ordinal position of a character (between 1 and 24), and a ground-truth label of that character in the image. The network used an occurrence loss (all characters present in the image) at the top of BU1 and a classification loss at the final output. This network was trained until learning the recognition task. Using this network as pre-trained weights rather than random weights, for the task of spatial relations, achieved performance of 94% using a single loss after 26 epochs, making end to end training successful (compared with the training failure noted above, without pre-training). Using two losses, it reached the same performance following 11 epochs, compared with 33 epochs starting with random weights. The initial learning is fast, reaching 87% accuracy after 3 epochs, compared with 4% accuracy after 3 epochs without pre-training.

## S3 The Persons data set

In addition to publicly available data sets we used the Persons and the Actors data sets described in this and the next section.

The Persons data set was generated using the graphics package http://avatarmaker.com/. Each person is constructed from two types of features - constant and variable. Constant features are the features that make this avatar unique, i.e. eyes, skin color and face shape, etc. The variable features, including haircut, glasses shape, mustache, and clothes, can change without affecting the avatar identity. Table 3 below lists the constant and variable features with the number of variations for each feature. The person images can be created in color, but in our experiments we use the grey-level version, as shown below.

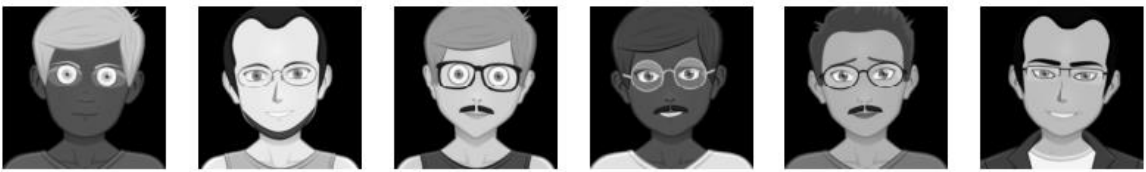

*Figure 32: Persons data set* Examples of different person images

The final images are constructed by placing a number of persons on a grayscale background image of size 448*224. To make person classification somewhat harder, in our images persons usually shared at least one constant feature with each of the other identities, so that a person could not be identified based on a single feature.

| Parameter | Face | Skin color | Lips | Lips color | nose | ears | eye front | iris | eyebrows | iris color | hair color | eyebrows color | beard color | beard |
|---|---|---|---|---|---|---|---|---|---|---|---|---|---|---|
| Variations | 15 | 20 | 15 | 20 | 15 | 7 | 15 | 10 | 15 | 20 | 20 | 20 | 20 | 13 |

| Parameter | Face Tilt | Clothes | Glasses | Hair | Mustache |
|---|---|---|---|---|---|
| Variations | 2 | 5 | 8 | 3 | 2 |

Table 3: Persons data set Top: constant parameters. Bottom: Variable parameters

## S4 The Actors data set

The Actors dataset domain consists of computer-generated scenes with human figures ('actors') objects and tools, created using Makehuman, an open-source tool for character making: http://www.makehumancommunity.org/). The actors we use come from four classes (woman, man, girl, boy), can have different poses, and can be designed with different properties. There is a broad range of possibilities, and in the main experiments we included four different outfits, optional sunglasses, optional hat, two hairstyles, and possibly carrying an object (backpack, shoulder bag), or holding an object (paper bag, bat, bow, hammer, tennis racket, signal, sword). Different objects of different types and colors can be placed in the scenery (bench, chair, trashcan, streetlight, and tree).

We defined several relations between actors in the scenes, which are annotated for use during training. For example, two actors may be placed in a 'facing' relation, and the model is trained to recognize 'facing' relations in novel images. Relations we used in the main experiments included spatial relations (left, right, in-front, behind), 'facing', 'touching', 'closest' (to another person or an object) and 'obstructing' (blocking the path of a second actor, within a given distance). An actor can interact with an object via a 'holding' relation, and relations between objects include the spatial relations and the 'on' relation. Images were generated by sampling a scene graph, i.e., sampling the objects and their properties, locations in the scene, and relations between them. The images are rendered using Blender (https://www.blender.org/), with given camera and lights locations. The generated images have annotations of all components, and have complete instance segmentation maps. A summary of the scene components, their properties and relations, as used in our experiments is given in Table 4 (Actor data set) below. Examples of scenes generated in Actors data set are illustrated in Figure 33 (Scene examples).

Finally, we list below briefly data on training and accuracy obtained at the output. The accuracy includes the average segmentation accuracy (measured by IoU, intersection-over-union), and average classification accuracy (of components, properties and relations).

Number of images used: Train: 12,000, Test: 3,353.
*Expansion network:*
Number of examples (an 'example' is a result of a TD instruction applied to an image, each scene contains multiple examples): Train: 638,825, Test: 176,596.
Results (train/validation): IOU at TD: 0.626/0.59, accuracy at BU2: 0.855/0.823
*Elaboration network:*
Number of examples: Train: 2,484,781, Test: 679,424
Results (train/validation): IOU at TD: 0.968/0.966, accuracy at BU2: 0.912/0.899

| Objects type | classes | Properties | Interacting relation |
|---|---|---|---|
| Humans | Man, woman, girl, boy | Clothes, hair, sunglasses, hat | Spatial, closest obstructing, facing, touching, holding |
| Scene objects | Bench, chair, trash can, streetlight, tree | Size, color | Spatial, closest obstructing, on |
| Held objects | Hammer, handbag, sword, racket, signal | Size, color | Holding, on |

| Relation | Ref object location $\vec{r_1}$ | Ref object Rotation $\theta_1$ | Related object location $\vec{r_2}$ | Related object Rotation $\theta_2$ | Other conditions |
|---|---|---|---|---|---|
| Obstructing | random | Random within visibility range | Random distance r within (0.6,3.5) from $\vec{r_1}$ in direction $\theta_1$. | within range that is visible to camera | |
| Facing | random | Random within visibility range s.t $\theta_1 + \pi$ also visible | r = average shoulder width of the ref, and the related object direction $\theta_1 \pm \frac{\pi}{2}$. | $\theta_2 = \theta_1 + \pi$ | |
| Touching | random | Random within visibility range | r within (0.4,3) from $\vec{r_1}$ in direction $\theta_1$. | $\theta_2 = \theta_1$ | The arms of both actors in touching position |
| Right of, left of, front of, behind, closest | random | Random in range visible to camera | Random | Random in range visible to camera | Relation calculated from location |

*Table 4: Actor data set* The table lists the scene components (people and objects), properties, and relations between components. 'Within visibility range' means that the person's rotation in the scene was constraint to make the face region visible.

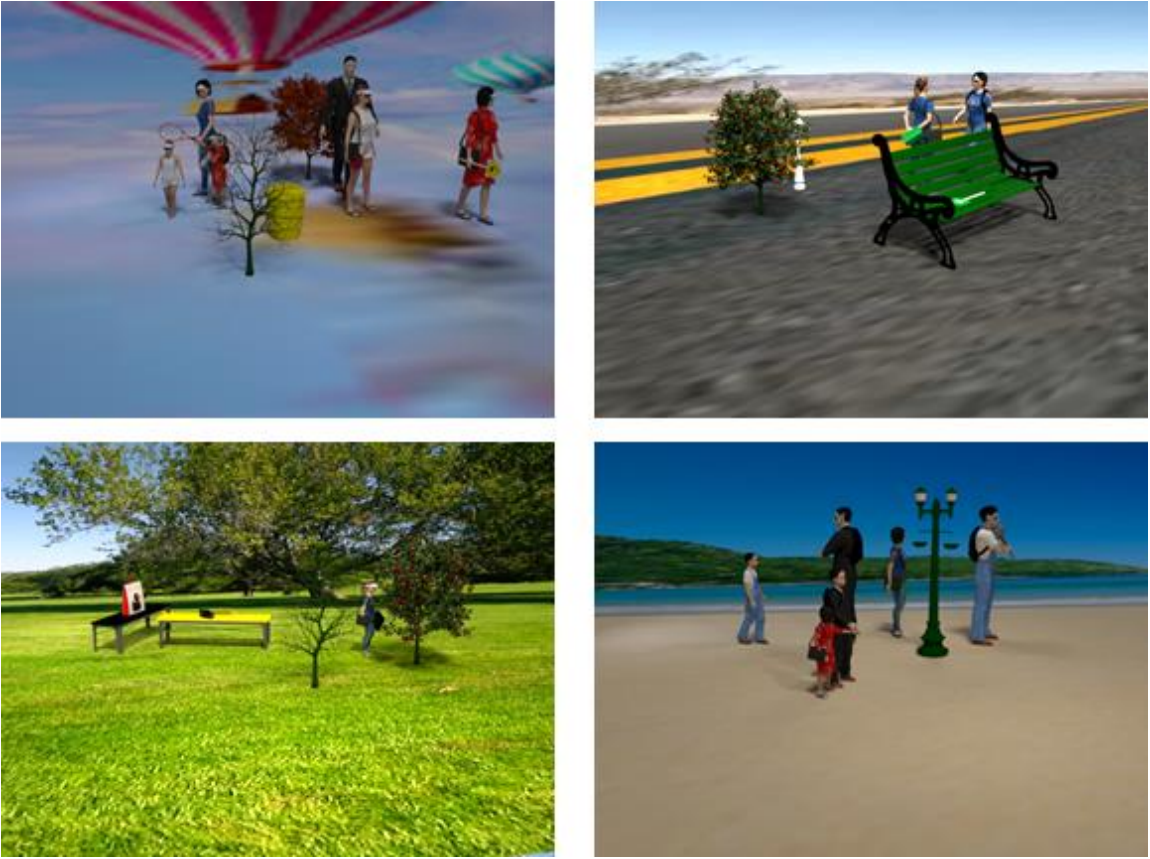

*Figure 33: Scene examples* Examples illustrating people, scene objects, and held objects used in the Actors data set.

## S5 Classification by location

In recognizing objects in a multi-object image (section 3.2 Objects, Figure 8b), a popular approach (e.g. (Ren *et al.*, 2015; Redmon *et al.*, 2016) is to detect and classify all of the objects in a given image. The approach here is to recognize a single selected object, using the BU-TD model, where the location of the object is specified by a TD instruction. We compared several alternatives for providing the selected location to the network. In the first comparison below, the instruction was given by a coarse localization map (stride 32) as an input for the TD stream. An alternative was to provide the selected location as a spatial map supplied in a BU manner together with the input image. In both cases, the object class is produced as an output at the end of the second BU stream.

We used a ResNet-50 backbone in both BU streams. All models were trained on COCO train2017 (118,000 images) and evaluated on COCO val2017 (5,000 images). The model was trained for 80,000 iterations with a classification loss at the end of the second BU stream. We compared the results with Faster-RCNN results, using its two main variants (termed C4 and FPN) using standard acc#1 and acc#5 as the classification metrics. Results are shown in Figure 34a below. Results are comparable to the accuracy of the highly optimized Faster-RCNN model.

Classification accuracy was also compared with the BU-TD model, when the location information is supplied as a spatial map supplied in a BU manner together with the input image. Results in Figure 34b as a function of the number of iterations used in training, show a significantly faster convergence of the instruction provided in a TD manner.

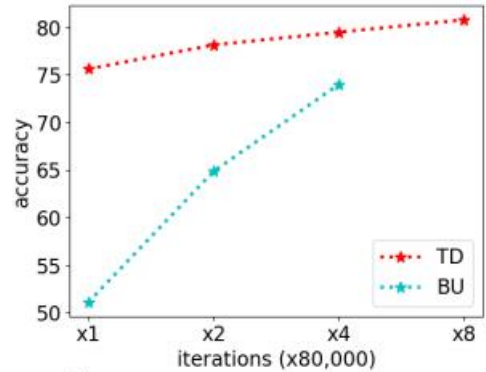

| | | acc#1 | acc#5 |
|---|---|---|---|
| Faster-RCNN | FPN | 83.167 | 96.752 |
| Faster-RCNN | C4 | 80.754 | 96.004 |
| Counter-Stream | TD | 80.919 | 96.810 |

a

b

*Figure 34: Classification by location* a. Classification accuracy of Faster-RCNN and the BU-TD counter-streams model using classification by location. b. Classification accuracy of the BU-TD model, when the location information is supplied in a TD manner (TD in the figure), or as a spatial map supplied in a BU manner together with the input image.

We also compared the results of the BU-TD models with other alternatives methods of specifying the location of the selected object: location of its central point, coordinates of the bounding box, an object mask. Asymptotic results were similar, but the use of a the TD map were highest, and learning by TD instruction was faster than the BU instruction.

## S6 Spatial relations

An example of training spatial relations is using images of multiple characters from the EMNIST data set (Cohen *et al.*, 2017), shown in Figure 35a below. Images of size 224*112 contained 24 characters, from a total set of 29 digits and letters. The characters in the input image were non-repeating. The network, shown in Figure 34b below, received two inputs – the input image, and a top-down instruction such as: <right-of, 1>, where the correct answer is A. The network is composed of BU1, TD and BU2 streams, each composed of 6 layers, with shared weights between BU1 and BU2.

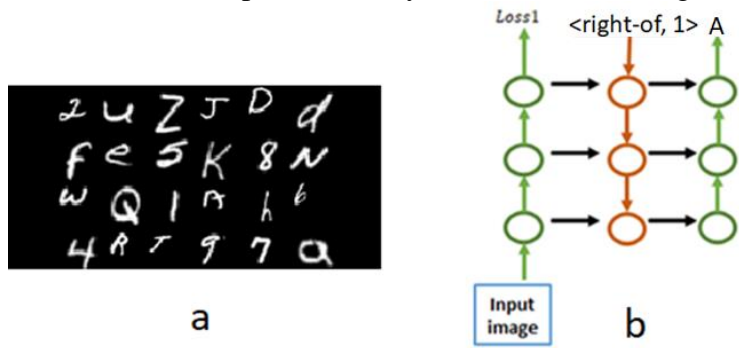

a

b

*Figure 35: EMNIST spatial relations* a. An input image, size 224*112. b. Schematic structure of the network, with TD instruction.

The network was trained from initial random weights, with two losses. At the top of BU1, the net was trained to identify all the characters present in the image, with binary cross entropy loss. The second loss was at the final output, producing a single character, using cross entropy loss. The training set included 10,000 images. During a single epoch, each image was shown 10 times with 10 different TD instructions, for a total of 100,000 single-character outputs.

Under these conditions, performance asymptoted at 96% accuracy at 39 training epochs. In comparison, using an unguided version of the same network (Section 5.2 (Combinatorial Generalization)) reached asymptotic performance of 48% following 537 epochs. (These results are for the non-generalization test. Combinatorial generalization results are discussed in Supplementary S10.)

*Non-referring spatial relations:*

Section 3.3 (referring relations) discussed examples of relation instructions using either a single object, in what was called a referring form, or using two object-arguments. An additional example of training the BU-TD model on relations by specifying the two relevant object, is the relation Occlude(x,y).

Using the Person data set, the instruction was given in the form of <occluding, (person-1 id, person-2 id)>, and the task is to learn the occlusion relationship between the two – whether the first occludes the second, or the first being occluded by the second, or there is no occlusion between the two. The output is therefore a classification into three classes. The dataset included 4800 training images. Each image had four persons, of which one pair participated in an occlusion relation. For each image, the training used three queries, where each time we select two of the persons. We adjusted the dataset such that half of the queries had an occlusion relation and half did not. The model used three losses: an occurrence loss (i.e., which persons appeared in the image) at the end of BU1, a final task loss at the end of BU2, and a TD segmentation loss (segmentation of the two persons in the instruction). The model obtained 99% task accuracy. Examples are shown in Figure 36 (Occlude(x,y)) below.

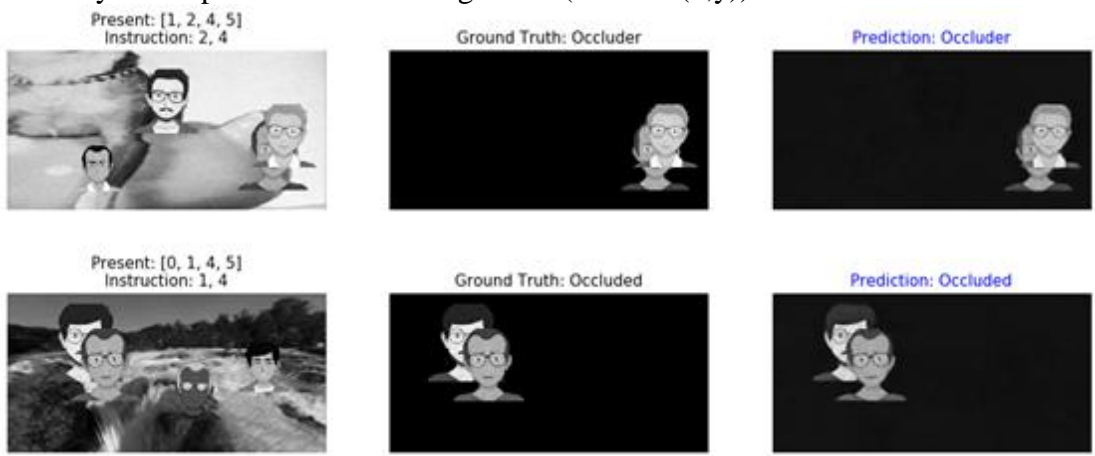

*Figure 36: Occlude(x,y)* The task is to compute the occlusion relation between person-1 and person-2: whether person-1 occludes person-2 ('occluder'), occluded by person-2, or no occlusion relations holds. Left: input images and the persons in the TD instruction, middle: segmentation produced by the TD stream, right: output prediction.

## S7 Extracting full scene structure

As described briefly in Section 4.1 (Extracting scene descriptions), 'full structure' means extracting all the scene components, their properties and relations. In analyzing real natural scenes, extracting their full structure is useful to examine the ability of the model to extract the full structure of scenes under controlled conditions, where the full structure is fully known. In this supplementary, we describe in more detail the algorithm used for extracting the full scene structure applied to the Actors data set. We first describe below the two networks used by the model, followed by the procedure for extracting the full scene structure.

The scene structure is extracted by the combination of two BU-TD networks: an 'expansion network' and an 'elaboration network'. The 'expansion network' extracts objects according to an instruction that can be a referring relation, or the 'extract-next' instruction, for the next closest 'main' item (a person or a scene object) to the camera. The 'elaboration network' recognizes a particular property/relation specified by the instruction.

**Expansion network**: This network is trained to find a person or an object based on its relation with a reference scene component, segment it and recognize its class (as described in Section 3.3 (Relations)). The referring instruction can be a binary relation, such as <facing, person-1> discussed above, which segments and classifies the person in the image facing person-1. The argument (person-1) is given to the net by its segmentation map, supplied together with the input image. The referring instruction can also be with respect to a set of objects and persons, rather than a single one. In particular, this is used to add a new person or object to the scene description, based e.g. on its distance from the camera, using the <extract-next> instruction. The argument to the <extract-next> instruction is given by a segmentation map that includes the persons and objects extracted so far from the image.

To produce segmentation maps, the top-down stream produces a score for each image pixel at the end of the TD stream, representing its probability of belonging to the target object, and the pixels that exceed a threshold are selected. Some post processing is carried out to produce the final object, to remove noise and redundant areas. To filter out small noisy candidates, morphology operations (erosion followed by dilation) are performed. When the selected pixels form multiple object candidates (unconnected regions), we filter out candidates, based mainly on their average pixel score.

**Elaboration network**: This network is trained to recognize a selected property of a specific item according to the TD instruction. The same network is also trained for relations of the form R(x,y), where the input is composed of two objects and a relation, and the output is the value of the relation (which can be for instance true/false or a selection between two complementary relations e.g. front/behind/none). The input arguments are provided by two segmentation maps that segment the two participating items in the relation (each map corresponds to one item). Technically, for property classification, one segmentation map is for the queried item and the second one is irrelevant (all zeros). The top-down stream is trained to provide the segmentation of the participating objects (one segmentation map).

The first step of obtaining full structure is to extract all scene components (objects and persons) by invoking the expansion network until 'no component' is returned, and storing them in an array. The second step is querying the properties for all components using the 'elaboration network' and updating the results for the corresponding component in the array. The particular set of property instructions is determined according to the item type (e.g. 'clothes' instructions is activated only for persons).

The final step is getting the relations between the components (for the spatial relation we use 'right' and 'behind' and exclude their complementary relations 'front' and 'left'). For each component, the 'expansion network' is activated multiple times, each time with a different relation instruction. The particular set of relations is determined according to the component type (e.g. 'facing' instruction is activated only for persons). An 'auxiliary-object' (e.g. hammer, hand-bag) is retrieved only by an instruction of a corresponding relation ('holding' or 'on') and not by the 'extract-next' instruction. The retrieved 'auxiliary-object' is added to the array and its properties are queried (using the elaboration network), and added as well. For other relations (with 'main' components), the retrieved component should be one of the components represented in the array. The component in the array with maximal IOU overlap is selected. If the maximal IOU is below a threshold, the new component is added to the array. A schematic 'pseudo-code' description of the algorithm is given below. Note that the functions *get_item_by_flag* and *get_prop_by_flag* are wrapper functions that invoke the expansion network and elaboration network, respectively.

---

**Algorithm 1:** Image full structure extraction using counter stream estimators

---

**Input:** image
**Result:** image graph
initialization: $image\_graph = [], item\_class = 0, items\_count = 0, scene\_mask =$ zeros;

**begin**
  **while** $item\_class$ is $\neg None \wedge items\_count < MAX\_ITEMS$ **do**
    $[item\_class, pred\_mask] = get\_item\_by\_flag(image, scene\_mask, flag = 'next\_item')^a$ ;
    $items\_count = items\_count + 1$ ;
    **if** $item\_class$ is $\neg None$ **then**
      $scene\_mask = scene\_mask \vee pred\_mask$ ;
      $overlap\_ind = get\_item\_ind\_by\_overlap(pred\_mask, image\_graph)$ ;
      **if** $overlap\_ind$ is $None$ **then**
        $item['class'] = item\_class$ ;
        $item['mask'] = pred\_mask$ ;
        $item['child\_nodes'] = [], item['child\_rels'] = []$ ;
        $item['parent\_nodes'] = [], item['parent\_rels'] = []$ ;
        **for** $prop\_type$ in $\mathbf{prop\_list}^b$ **do**
          $item[prop\_type] = get\_prop\_by\_flag(image, pred\_mask, flag = prop\_type)^c$ ;
        **end**
        $image\_graph.append(item)$ ;
      **end**
    **end**
  **end**
  **for** $item\_ind$ in $length(image\_graph)$ **do**
    **for** $rel$ in $\mathbf{rels\_list}^d$ **do**
      $[rel\_item\_class, rel\_pred\_mask] = get\_item\_by\_flag(image, image\_graph[item\_ind][mask], flag = rel)^a$ ;
      $child\_ind = get\_item\_ind\_by\_overlap(rel\_pred\_mask, image\_graph)$ ;
      **if** $rel$ in $tool\_rels \vee child\_ind$ is $None$ **then**
        $new\_item['class'] = rel\_item\_class$ ;
        $new\_item['mask'] = rel\_pred\_mask$ ;
        $new\_item['parent\_nodes'] = [], new\_item['parent\_rels'] = []$ ;
        **for** $prop\_type$ in $\mathbf{prop\_list}^b$ **do**
          $new\_item[prop\_type] = get\_prop\_by\_flag(image, rel\_pred\_mask, flag = prop\_type)^c$ ;
        **end**
        $child\_ind = length(image\_graph)$ ;
        $image\_graph.append(new\_item)$ ;
      **end**
      $image\_graph[child\_ind]['parent\_nodes'].append(item\_ind)$ ;
      $image\_graph[child\_ind]['parent\_rels'].append(rel)$ ;
      $image\_graph[parent\_ind]['child\_nodes'].append(child\_ind)$ ;
      $image\_graph[parent\_ind]['child\_rels'].append(rel)$ ;
    **end**
  **end**
**end**

---

$^a get\_item\_by\_flag$ function invokes the 'expansion network'
$^b$ property list is adopted according to item type (e.g. 'clothes' is relevant only for persons)
$^c get\_prop\_by\_flag$ function invokes the 'elaboration network'
$^d$ relation list is adopted according to item type (e.g. 'facing' is relevant only for persons)

---

## S8 Guided extraction of scene structure

In the guided interpretation, we find a grounding of a query structure in the image by executing a sequence of BU-TD tasks guided by the given structure. In this section, we only describe briefly the data structures used to represent the extracted structure and its grounding in the image. The process of guiding the BU-TD network to extract the structure of interest is described in the next supplementary (Supplementary S9 (Selecting the next instruction)).

The procedure maintains and updates two data structures:

(1) An array of objects (including persons) and their details, which are updated according to the output of the BU-TD activations. This represents the retrieved structure grounded in the query image.

(2) A query structure in the form of a graph, where nodes represent objects and edges represent relations, which is used to guide the procedure. Each node includes a pointer to a corresponding object from the array of objects that is currently a valid candidate for this node. We update these pointers according to tests' results.

Once the query processing is finished, the array of objects includes all the detected objects and their corresponding information as detected during the 'grounding' procedure (including indices of related objects). This array, which is essentially the extracted structure (graph), is a subset of the 'full structure' and a superset of the query structure in case of a successful query validation. The array can be used for 'follow up' inquiries, where only the missing information will be computed.

## S9 Selecting the next TD instruction: composing visual routines

The procedure used to produce the TD instructions is a recursive process that handles one node at a time (starting at a root node if the graph has a root node, and an arbitrary node otherwise). As above, the full process is carried out by the combination of two network models, described above in Supplementary S7 (Extracting full scene structure). The first ('expansion network'), is responsible for extending the graph, by adding a new item (person or object) detected in the image. The second (elaboration network) is used for elaborating the existing graph, by extracting additional properties and certain relations according to the required structure. We outline next the procedure guiding the model to extract the target structure, which builds upon the algorithm described in detail in (Vatashsky and Ullman, 2020).

The procedure extracts persons and objects using the expansion network. In the Actors data set the persons and scene objects (e.g. trees and benches) are extracted one by one using the <extract-next> instruction. The algorithm proceeds along the nodes in the target structure, and it guides the model to extract from the image people and objects with their properties and relations that will correspond to the structure described by the target graph. If the class of the extracted person corresponds to the class required by the current node in the graph (e.g. 'girl'), the process checks the requirements of the node in terms of properties and relations. The relevant properties are extracted and validated using the appropriate TD property instructions, and the relevant relations are extracted and validated using the expansion network, which retrieves the persons and objects required by the relations in the graph (e.g. 'facing'). Spatial relation instructions (e.g. 'right-of', 'behind') extract people and scene objects, and relations such as 'holding' or 'on' are used to extract tools and objects held by people and placed on other objects. Property extraction by the elaboration network is used either to validate a property in the target graph (e.g., looking for a red object), or retrieving a property (e.g., finding the color of the object on the table). The retrieved information is saved and updated in an array where each entry represents data of one object (including a person). The retrieved objects that fulfill the requirements are paired with the corresponding objects in the target graph, so that subsequent tests will be applied to the correct objects. The number of required objects is set according to quantifiers (e.g. 'all', 'three'), or by the need to evaluate a property that depends on the entire object set (e.g. 'how many'). Once all the node's tests are completed, the process repeats for the next node, determined by a depth-first (DFS) traversal. After the node and the following nodes in the recursion are tested and validated in the image, tests are applied, if needed, to verify the so-called node's set requirements. Examples for such tests are counting the objects in the set and quantity comparisons. The recursive process is terminated either when the full graph is grounded in the image as required, or when no alternatives are left to check, which may occur already after a partial evaluation of the image, e.g. no objects of a required class were detected. Pseudo-code description of the algorithm for selecting the next TD instruction and extracting a structure of interest is given below.

---

**Algorithm 2:** Guided scene structure extraction using counter stream estimators

---

**Input:** query_graph, image
**Result:** query answer, retrieved structure graph
initialization: $workMem['current\_node'] = first\_parent\_node$;
**Run**$[success, answer] = getGraphAnswer$ :
  **begin**
    $current\_node = workMem['current\_node']$ ;
    Node parameters: $\boldsymbol{p}$: properties, $\boldsymbol{r}$: relations, $f$: property type, $g$: property of a set;
    $last\_item, no\_items, no\_child\_items$ =False; $mask$ =zeros;
    **while** $\neg no\_items \wedge \neg last\_item$ **do**
        **if** ***child_node*** **then**
            $item = workMem['item']$ ;
            $last\_item$ =True ;
        **else if** $is\_super\_node^a$ **then**
            $[item, last\_item] = get\_item\_from\_saved\_sub\_nodes$
        **else if** $exist(saved\_detected\_items)$ **then**
            $[item, last\_item] = get\_item\_from\_saved\_dtected$
        **else**
            $[item, no\_items] = get\_item\_by\_flag(image, mask, flag = 'next\_item')^b$ ;
            $mask = mask \vee item['seg']$ ;
        **end**
        **if** $\neg empty(\boldsymbol{p})$ **then**
            **for** $p$ in $\boldsymbol{p}$ **do**
                $\hat{p} = get\_prop\_by\_flag(image, item['seg'], flag = f_p)^c$ ;
                $success = \hat{p} == p$ ;
                **if** $\neg$ success **then** break **end**
            **end**
            **if** $\neg$ success **then** continue **end**
        **end**
        **if** $\neg empty(f)$ **then** $answer = get\_prop\_by\_flag(image, item['seg'], flag = f)^c$ **end**
        **if** $empty(\boldsymbol{r})$ **then**
            **if** $exist(next\_parent\_node)$ **then**
                $workMem['current\_node'] = next\_parent\_node$;
                Run $[success, answer] = getGraphAnswer$;
            **end**
        **else**
            **for** $r$ in $\boldsymbol{r}$ **do**
                $child\_mask = item['seg']$;
                **while** $\neg no\_child\_items$ **do**
                    **if** $empty(item[r\_item])$ **then**
                        $[item[r\_item], no\_child\_items, answer] = get\_item\_by\_flag(image, child\_mask, flag = r)^b$;
                    **end**
                    $child\_mask = item[r\_item]['seg']$ ;
                    $sucess = \neg no\_child\_items$ ;
                    **if** $success$ **then**
                        $workMem['current\_node'] = next\_node^d$;
                        $workMem['item'] = child\_item$;
                        Run $[success, answer] = getGraphAnswer$;
                        **if** success $\wedge$ (#success_child_items == #required_child_items$^e$) **then** break **else** success = False **end**
                    **end**
                **end**
                **if** $\neg$ success **then** break **end**
            **end**
        **end**
        **if** success $\wedge$ (#success_items == #required_items$^e$) **then** break **end**
    **end**
    **if** success $\wedge \neg$ empty(g) **then** $answer = g(valid\_items)$ **end**
    **if** success $\wedge$ comp_num_en $\wedge$ is_checked(comp_node) **then** $answer = comp\_num^f(valid\_items, comp\_items, comp\_type)$; **end**
    **if** $success \wedge is\_sub\_node$ **then** $save\_for\_super\_node(success\_items)$; **end**
    return $[success, answer]$
  **end**

---

$^a$A 'super node' includes other nodes ('sub nodes')
$^b get\_item\_by\_flag$ function invokes the 'expansion network'
$^c get\_prop\_by\_flag$ function invokes the 'elaboration network'
$^d$Either child node or next unvisited root node of a subgraph
$^e$According to quantifiers and other requirements
$^f$Compare number of valid items between nodes ('same', 'fewer', 'more')

# S10 Combinatorial generalization

In our combinatorial generalization tests we compared four models on two tasks: extracting Persons properties, and computing spatial relations (left-of, right-of) between EMNIST characters. The four models in these comparisons are illustrated schematically in Figure 37 (Comparing combinatorial models). The main comparison was between our BU-TD model (in a.), and the model labeled 'Unguided, Readout selection' (in b.). The two models are almost identical in structure, but, as explained in Section 5.2 (Combinatorial generalization), the crucial difference is that the BU-TD model is guided by an instruction, while the Readout selection is unguided. Unlike the BU-TD model, to provide a correct output to any selected instruction, the readout selection model must extract simultaneously all the properties of the persons in the image in a single pass. The readout selection does not determine which property to extract, but only selects the readout from the last layer.

The two other models are different versions of guided models. The BU instruction, Guided model (in c.) is identical to the BU-TD model, but the task-instruction is given in a BU manner, together with the input image. This is done by learning an embedding from the instruction to a vector of the same size of the image. The final model is a standard ResNet model (in d.), which is also provided with a task instruction at the input level, in a manner similar to the BU instruction, Guided model.

The ResNet has a similar structure, number of units and connections to the BU-TD model, but without the lateral connections between the BU and TD streams. It has the same number of layers as the BU-TD, but without shared parameters, therefore, the number of parameters is about 1.5 times larger. The other models have the same number of parameters, except for the size of the instruction embedding layer in the BU instruction, Guided model, which makes it larger than the embedding in the BU-TD and Readout instruction models.

We used the same losses on all models: the first one is an occurrence loss (i.e., which characters or persons are in the image) and the second is the task loss (character right/left-of or a person's property detail).

Similar to the BU-TD model, the models in c. and d. are guided by a task instruction to perform a selected task. In principle, they might be able to perform in a similar manner: the instruction at the input level in c. could be propagated along the BU stream without a change until it reached the top BU layer, and then proceed to the TD stream, similar to instruction in the BU-TD model. They were used here not to distinguish between guided and unguided models, but to make comparisons between different variations of guided models.

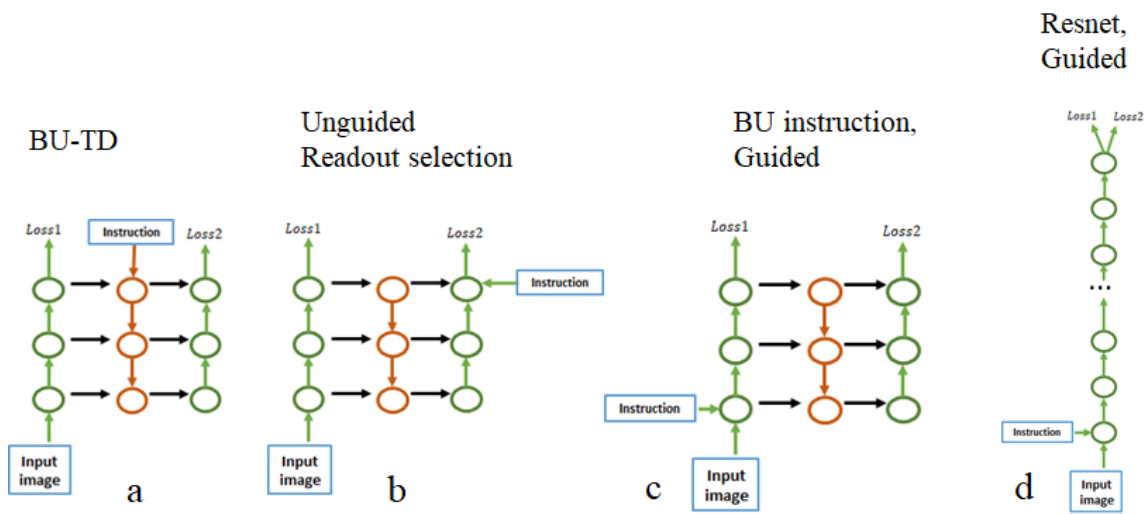

*Figure 37: Comparing combinatorial models* A simplified diagram of the models used in the comparisons; each circle stands for a layer. a. BU-TD model architecture. b. Unguided, Readout selection. c. BU instruction Guided. d. ResNet, Guided. The main comparison was between the BU-TD model (in a.), and the Unguided, Readout selection (in b.). The two models are almost identical in structure, but the BU-TD model is guided by an instruction, and the Readout selection is unguided.

The embedding size for BU-TD and for Bottom Up (Readout) models is of size 256 for the EMNIST experiments and 512 for the Persons experiments. For the BU guided and ResNet guided models (where it depends on the image size), the embedding is of size (W/2 x H/2) where W and H are the image width and height, and then reshaped and up-scaled to (WxH) size, and concatenated with the input image. For the EMNIST experiments the embedding size is 6,272 and for the Persons experiments 25,088.

### Details of the experiments

We performed combinatorial generalization experiments on two different tasks, EMNIST and Persons. For each task, we used two dataset sizes: a sufficient size and an extended size. The sufficient size is the smallest dataset for which BU-TD achieved an accuracy of at least 85%. In different experiments, we trained all four models under comparison on this sufficient set, and examined the performance obtained by the different models. In addition to the training set, we created a validation set, which had the same distribution as the training set (specifically, did not use the 'excluded pairs' of object-property or object-relation discussed in section 5.2 (Combinatorial Generalization)), and combinatorial test data, which focused on the excluded pairs, to test generalization to novel combinations. A similar scheme was used for the extended set experiments.

We performed extensive hyper-parameters search for all models with over 30 combinations of Optimizer: SGD (stochastic gradient descent) with momentum and ADAM (Kingma and Ba, 2015), Learning rate (6 values: 0.0001, 0.001, 0.002, 0.05, 0.1, 0.2), Batch size (10, 32, 48) and Weight Decay (0.0001, 0.0002). The search was 'fair' in the sense that the same set of hyper-parameters was used for all the models participating in the comparisons. If models in the comparisons failed to reach the performance of the BU-TD model, we further searched for such models additional training alternatives, including Batch Normalization (Ioffe and Szegedy, 2015) and Group Normalization (Wu and He, 2020), searching over group normalization hyper parameters. We also tried different optimizers such as the Lookahead optimizer (Zhang *et al.*, 2019), AdamW and Adamax (Loshchilov and Hutter, 2019).

### Combinatorial generalization – Persons data set

For the Persons dataset we have two persons in each image with 5 properties each. Therefore, we generated several training examples per image, one for each of the person-property pairs. The Sufficient dataset had 1600 training images from which we generate 22,400 training examples. We had additional 800 (non-combinatorial) validation examples and 1176 test (combinatorial) examples. In the Extended dataset we had 4,800 training images and 67,144 examples, with additional 1198 (non-combinatorial) validation examples and 1176 test examples.

*Results*: For each model, we report the accuracy, as well as the number of epochs, and number of shown examples (the number of examples in the dataset times the number of epochs; in parenthesis, in thousands.). Accuracy is measured after reaching the following convergence criterion: we report accuracy A at epoch E where the accuracies in the subsequent 50 epochs increase by at most 2%.
In Table 5 below we report the results of the Persons experiments.

|  | BU-TD | Unguided Readout selection | BU Instruction, Guided | ResNet, Guided |
|---|---|---|---|---|
| Sufficient data, Non generalization | 89% 187 (4,188) | 68% 48 (1,075) | 55% 13 (291) | 80% 263 (5,891) |
| Sufficient data, Generalization | 89% 181 (4,054) | 31% 71 (1,590) | 22% 47 (1,052) | 65% 573 (12,835) |
| Extended data, Non generalization | 97% 53 (3,558) | 96% 108 (7,251) | 98% 91 (6,110) | 98% 100 (6,714) |
| Extended data, Generalization | 93% 71 (4,767) | 29% 213 (14,301) | 88% 94 (6,311) | 96% 121 (8,124) |

*Table 5: Combinatorial generalization in the Persons data set* For each test, accuracy is shown on the top line; number of epochs, and number of shown examples (in parentheses) are shown on the second line. The main result is that the unguided model (Readout selection) has poor combinatorial generalization compared with the BU-TD model. In contrast, the unguided model reaches comparable accuracy for non-combinatorial generalization with extended data. The two additional guided models reach high combinatorial generalization when trained with the extended data set.

### Combinatorial generalization – EMNIST data set

For the EMNIST Sufficient dataset, we generated 10 examples per image, resulting in 100,000 training examples, and additional 500 validation examples and 500 test examples. In the Extended dataset there are 24 characters in each image. We generated 20 examples for the right-of task and 20 for the left-of for a total of 40 examples per image. We have 10,000 images resulting in 400,000 training examples, and additional 500 validation examples and 500 test examples.

We report the results in Table 6 below.

|  | BU-TD | Unguided Readout selection | BU Instruction, Guided | ResNet, Guided |
|---|---|---|---|---|
| Sufficient data, Non generalization | 96% 39 (3,900) | 48% 537 (53,700) | 5% 2 (200) | 1% 1 (100) |
| Sufficient data, Generalization | 95% 42 (4,200) | 14% 635 (63,500) | 8% 81 (8,100) | 4% 1 (100) |
| Extended data, Non generalization | 95% 18 (7,200) | 75% 294 (117,600) | 95% 136 (54,400) | 95% 15 (6,000) |
| Extended data, Generalization | 91% 15 (6,000) | 10% 135 (54,000) | 96% 145 (58,000) | 95% 17 (6,800) |

*Table 6: Combinatorial generalization in the EMNIST-24 data set* For each test, accuracy is shown on the top line; number of epochs, and number of shown examples (in parentheses) are shown on the second line. The main result is that the unguided model (Readout selection) has poor combinatorial generalization compared with the BU-TD model. The two additional guided models reach high accuracy in combinatorial generalization when trained with the extended data set.

The pattern of results is similar on the two data sets. The main comparison is between the BU-TD model, which is guided by a task-instruction, and the similar, but unguided BU model. The main result is that the unguided BU model shows severely limited combinatorial generalization. The other forms of guided models also achieve high accuracy in combinatorial generalization, but training can take longer compared with the BU-TD model.

Finally, we also conducted generalization experiments on a simpler task, called EMNIST-6, with 6 characters per image. The purpose of this experiment was to test whether combinatorial generalization can start to emerge in an unguided model, well above chance when the task is simplified and the training is sufficiently extended. We show below results for an EMNIST task similar to the one above, when the number of characters was reduced from 24 to 6. For the EMNIST-6 Sufficient dataset we generated 5 examples per image (excluding the rightmost character). We used 2,000 images resulting in 10,000 training examples, along with additional 500 validation examples and 500 test examples. The extended dataset had 10,000 training images resulting in 50,000 training examples, with additional 500 validation examples and 500 test examples. As seen in Table 7, the unguided model can achieve combinatorial generalization when the task is simple and the dataset is large enough (although with a lower accuracy).

|  | BU-TD | Unguided Readout selection | BU Instruction, Guided | ResNet, Guided |
|---|---|---|---|---|
| Sufficient data, Non generalization | 84% 95 (950) | 8% 15 (150) | 8% 15 (150) | 81% 815 (8,150) |
| Sufficient data, Generalization | 89% 142 (1,420) | 13% 194 (1,940) | 18% 36 (360) | 89% 923 (9,230) |
| Extended data, Non generalization | 95% 50 (2,500) | 80% 203 (10,150) | 95% 38 (1,900) | 95% 25 (1,250) |
| Extended data, Generalization | 94% 43 (2,150) | 77% 551 (27,550) | 95% 45 (2,250) | 96% 56 (2,800) |

*Table 7: Combinatorial generalization in the EMNIST-6 data set* For each test, accuracy is shown on the top line; number of epochs, and number of shown examples (in parentheses) are shown on the second line.

For the unguided model, we also tested a parallel-learning model, instead of the Readout selection. We trained EMINST-6 with a standard BU model, predicting all right-neighbors together. This was done with a 'matrix', or a table representation, where M(i,j) means that character j is the right-neighbor of character i. In this representation, the output units representing the pairs excluded from training, were never activated during training, and therefore activation of the correct output units cannot be expected. We tested whether the correct answers were in fact encoded in the output layer of the network by training a readout from this last layer. The readout training took place also after the full training of the main network. Using this readout network we found that the parallel-learning version reached the same generalization accuracy as the sequential selective readout for the right-of task. The comparisons indicate that the selective readout network is equivalent to a network trained to produce in parallel the predictions for all the different tasks, but it has the advantage that results for the pairs excluded during training can be read out in a manner that is not possible in the more standard table representation or in a multi-branch architecture.

The results of combinatorial generalization across the different tasks show that combinatorial generalization is severely limited in the unguided model compared with the guided models. Combinatorial generalization became possible in the spatial relations task when the number or characters was reduced from 24 to 6. Keeping the number of characters at 6, but making the task more complex reduced combinatorial generalization in the unguided model compared with the guided model. For example adding the left-of relation (so that the task is either right-of or left-of a selected character), combinatorial generalization of the unguided model was reduced to 77% (vs. 95% of the BU-TD). Increasing the proportion of excluded pairs to 63% caused the unguided model to drop sharply to 3% (vs 82% of the BU-TD).

From all the experiments above, we conclude that combinatorial generalization in the unguided model is not impossible, but is severely limited compared with the guided models. The gap between the guided and unguided models increases when the task becomes more complex (e.g., additional relations, a larger number of objects), or when the fraction of combinations excluded during training decreases. For the task of scene interpretation, when the number of trained combinations can become only a fraction of all possible combinations, guided models may be the only feasible alternative.